\documentclass{article} %

\usepackage[dvipsnames]{xcolor}
\usepackage[accepted]{icml2021}

\usepackage{url}

\usepackage{array}
    \newcolumntype{C}{>{$}c<{$}}
    \newcolumntype{L}{>{$}l<{$}}
    \newcolumntype{R}{>{$}r<{$}}

\usepackage{microtype}
\usepackage{graphicx}
\usepackage{subfigure}
\usepackage{booktabs} %
\usepackage{amsmath, latexsym}
\usepackage{amsfonts}
\usepackage{amssymb}
\usepackage{bm}
\usepackage{mathtools}
\usepackage[normalem]{ulem}
\usepackage[most]{tcolorbox}
\usepackage{placeins}
\usepackage{enumitem}
\usepackage{bold-extra}
\usepackage{chngcntr}

\usepackage[T1]{fontenc}

\usepackage{listings}
\definecolor{backcolour}{rgb}{0.95,0.95,0.92}
\definecolor{codegreen}{rgb}{0,0.6,0}

\lstdefinestyle{myStyle}{
    backgroundcolor=\color{backcolour},   
    commentstyle=\color{codegreen},
    basicstyle=\ttfamily\footnotesize,
    breakatwhitespace=false,         
    breaklines=true,                 
    keepspaces=true,                 
    numbers=left,       
    numbersep=5pt,                  
    showspaces=false,                
    showstringspaces=false,
    showtabs=false,                  
    tabsize=2,
    keywordstyle=\color{blue},       %
}

\usepackage{xspace}
\usepackage{makecell}

\usepackage{newfloat}
\DeclareFloatingEnvironment{eqn}

\usepackage{standalone}
\usepackage{tikz}
\usepackage{pgfplots}
\usepackage{pgf}
\usetikzlibrary{calc}
\usetikzlibrary{positioning}
\usetikzlibrary{angles,quotes}
\usetikzlibrary{backgrounds}
\usetikzlibrary{external}
\usetikzlibrary{fit}
\usetikzlibrary{arrows}
\usetikzlibrary{arrows.meta}
\usetikzlibrary{shapes.symbols}
\usetikzlibrary{shadings}
\usetikzlibrary{shapes}
\usetikzlibrary{fadings}
\usetikzlibrary{bayesnet}
\usetikzlibrary{matrix}
\usetikzlibrary{snakes}
\usetikzlibrary{decorations.pathmorphing, patterns}
\pgfplotsset{compat=1.15}

\renewcommand{\edge}[3][]{ %
	\foreach \x in {#2} { %
		\foreach \y in {#3} { %
			\path (\x) edge [->, >={stealth'},  line width=1pt,
			shorten <=0.75pt,
			shorten >=0.75pt,  #1] (\y) ;%
		} ;
	} ;
}

\definecolor{C1}{HTML}{1F77B4}
\definecolor{C2}{HTML}{FF7F0E}
\definecolor{C3}{HTML}{2CA02C}
\definecolor{C4}{HTML}{D62728}
\definecolor{C5}{HTML}{9467BD}
\colorlet{C1light}{C1!70!white}
\colorlet{C2light}{C2!70!white}
\colorlet{C3light}{C3!70!white}
\colorlet{C4light}{C4!70!white}
\colorlet{C5light}{C5!70!white}
\colorlet{C1vlight}{C1!20!white}
\colorlet{C2vlight}{C2!20!white}
\colorlet{C3vlight}{C3!20!white}
\colorlet{C4vlight}{C4!20!white}
\colorlet{C5vlight}{C5!20!white}
\colorlet{linkcolor}{violet}
\colorlet{citecolor}{RedOrange}  %
\colorlet{urlcolor}{Aquamarine}

\usepackage{caption}
\DeclareCaptionFont{textsc}{\small\bfseries\selectfont}
\usepackage{hyperref}
    \newcommand\myshade{85}
    \hypersetup{
      linkcolor  = linkcolor!\myshade!black,
      citecolor  = citecolor!\myshade!black,
      urlcolor   = urlcolor!\myshade!black,
      colorlinks = true,
    }

\usepackage[capitalise, nameinlink]{cleveref}
    \Crefname{table}{Tab.}{Tabs.}
    \Crefname{appendix}{App.}{Apps.}
    \Crefname{section}{Sec.}{Secs.}
    \Crefname{equation}{Eq.}{Eqs.}
    
\usepackage{defns}
\usepackage{dblfloatfix}

\tikzset{
    /pgf/decoration/amplitude = 0.1em,
    /pgf/decoration/segment length = 0.5em}
\usepackage{cuted}
\usepackage{flushend}

\allowdisplaybreaks

\begin{document}

\addtocontents{toc}{\protect\setcounter{tocdepth}{-1}}

\newcommand{\ourtitle}{Generalized Doubly-Reparameterized Gradient Estimators}
\icmltitlerunning{\ourtitle}

\twocolumn[

\icmltitle{\ourtitle}

\begin{icmlauthorlist}
\icmlauthor{Matthias Bauer}{dm}
\icmlauthor{Andriy Mnih}{dm}
\end{icmlauthorlist}
\icmlcorrespondingauthor{Matthias Bauer}{msbauer@deepmind.com}
\icmlcorrespondingauthor{Andriy Mnih}{andriy@deepmind.com}
\icmlaffiliation{dm}{DeepMind, London, UK} 

\vskip 0.3in
]

\printAffiliationsAndNotice{}

\begin{abstract}
Efficient low-variance gradient estimation enabled by the reparameterization trick (RT) has been essential to the success of variational autoencoders. Doubly-reparameterized gradients (\dregs) improve on the RT for multi-sample variational bounds by applying reparameterization a second time for an additional reduction in variance. Here, we develop two generalizations of the \dregs estimator and show that they can be used to train conditional and hierarchical VAEs on image modelling tasks more effectively. First, we extend the estimator to hierarchical models with several stochastic layers by showing how to treat additional score function terms due to the hierarchical variational posterior. We then generalize \dregs to score functions of arbitrary distributions instead of just those of the sampling distribution, which makes the estimator applicable to the parameters of the prior in addition to those of the posterior.
\end{abstract}

\section{Introduction}

In probabilistic machine learning we %
often optimize expectations of the form $\varL_{\vphi, \vtheta} = \expect{\distqz}{f_{\vphi, \vtheta}(\vz)}$ \wrt to their parameters, where $f_{\vphi, \vtheta}(\vz)$ is some objective function, and $\vphi$ and $\vtheta$ denote the parameters of the sampling distribution $\distqz$ and other (e.g.~model) parameters, respectively. In the case of the influential variational autoencoder (VAE, \citet{Kingma_vae}, \citet{Rezende2014_vae}), $\distqz$ is the variational posterior, $\vtheta$ denotes the model parameters, and $\varL_{\vphi, \vtheta}$ is typically either the ELBO \citep{Jordon1999_variational_methods,Blei2017_vi_review} or IWAE \citep{Burda2015_iwae} objective.

In most cases of interest, this expectation is intractable, and we estimate it and its gradients, $\grad{\vphi}{\varL}$ and $\grad{\vtheta}{\varL}$, using Monte Carlo samples $\vz \sim \distqz$. 
The resulting gradient estimators are characterized by their \emph{bias} and \emph{variance}. We usually prefer unbiased estimators as they tend to be better-behaved and are better understood. Lower variance is also preferable because it enables faster training by allowing using higher learning rates.

In this paper, we address gradient estimation for continuous variables in variational objectives. A naive implementation of $\grad{\vphi}{\varL}$ results in a \emph{score function}, or REINFORCE, estimator \citep{williams1992simple}, which tends to have high variance; however, if $\fz$ depends on $\vphi$ only through $\vz$, we can use  reparameterization \citep{Kingma_vae,Rezende2014_vae} to obtain an estimator with lower variance by replacing the score function estimator of the gradient with a \emph{pathwise estimator}. 

In variational inference, $\fz$ typically depends on $\vphi$ not only through $\vz$ but also through the value of the log density $\lqz$. %
Then, %
the gradient estimators still involve the score function $\grad{\vphi}{\lqz}$ despite using reparameterization. \citet{Roeder2017_stl} propose the \emph{sticking the landing} (\stl) estimator, which simply drops these score function terms to reduce variance. %
\citet{Tucker2019_dregs} show that \stl is biased in general, and introduce the \emph{doubly-reparameterized gradient} (\dregs) estimator for IWAE objectives, which again yields unbiased lower-variance gradient estimates. This is achieved by applying reparameterization a second time, targeting the remaining score function terms.

However, the \dregs estimator has two major limitations: 1) it only applies to latent variable models with a single latent layer; 2) it only applies in cases where the score function depends on the same parameters as the sampling distribution. In this work we address both limitations. Moreover, we show that for hierarchical models with several stochastic layers, gradients that look like pathwise gradients can actually contain additional score function gradients that are not doubly reparameterizable. Despite this, we show that we can still obtain a simple estimator with a sizable reduction in gradient variance for hierarchical IWAE objectives.

Our main contributions are:
\begin{itemize}[topsep=-2pt,itemsep=2pt,partopsep=0pt, parsep=0pt, leftmargin=*]
    \item We extend \dregs to hierarchical models;
    \item We introduce \gdregs, a generalization of \dregs to score functions that depend on a different distribution than the sampling distribution;
    \item We show how to implement all proposed gradient estimators using automatic differentiation frameworks;
    \item We evaluate the proposed \dregs and \gdregs estimators on several conditional and unconditional unsupervised learning problems and find that they outperform the regular IWAE estimator.
\end{itemize}

\section{Background}

In this work we are interested in computing gradients of variational
objectives of the form %
\begin{align}
    \label{eq:objective}
    	\varL_{\vphi, \vtheta} = \expect{\vz\sim\distqz}{f_{\vphi,\vtheta}(\vz)} %
\end{align}
\wrt the variational parameters $\vphi$ of the sampling distribution $\distqz$, and parameters $\vtheta$ of a second distribution $\distpz$, such as a learnable prior. 
Here $f_{\vphi,\vtheta}(\vz)$ is a general function of $\vz$ that can also explicitly depend on both $\distqz$ and $\distpz$. 
More precisely, we wish to compute
\begin{align}
    \gradtd{\vphi}{\varL_{\vphi, \vtheta}} \qquad \text{and} \qquad \gradtd{\vtheta}{\varL_{\vphi, \vtheta}},
    \label{eq:objective_gradients}
\end{align}
where $\gradtd{\ast}{}$ denotes the total derivative, which we explicitly distinguish from the partial derivative $\grad{\ast}{}$.

Arguably the simplest objectives of this form are the negative entropy $\varL^\text{ent}_{\vphi, \vtheta}=\expect{\vz\sim\distqz}{\log\distqz}$ and negative cross-entropy $\varL^\text{ce}_{\vphi, \vtheta}=\expect{\vz\sim\distqz}{\log\distpz}$.

\paragraph{Importance weighted autoencoders.} Another such objective is the importance weighted autoencoder (IWAE) bound \citep{Burda2015_iwae}. For a VAE with likelihood $\likelzl$, (learnable) prior $\distpz$, and variational posterior (or proposal) $\qzx$ the IWAE bound with $K$ importance weights $w_k = \tfrac{\pzk \pxzkl}{\qzkx}$ is given by
\begin{align}
    \liwae & = \expect{\vz_1, \dots, \vz_K \sim \qzx}{\log\left(\tfrac{1}{K}\textstyle\sum_{k=1}^K w_k\right)}.
    \label{eq:iwae}
\end{align}
\Cref{eq:iwae} reduces to the regular ELBO objective for $K=1$ %
\citep{Rezende2014_vae,Kingma_vae},
\begin{align}
    \varL^\text{ELBO}_{\vphi, \vtheta} & = \expect{\vz\sim\qzx}{\log\tfrac{\distpz\likelz}{\qzx}}.
\end{align}
\citet{Burda2015_iwae} showed that using multiple importance samples ($K>1$) provides the model with more flexibility to learn richer representations (fewer inactive units), and results in better log-likelihood estimates compared to VAEs trained with the single sample ELBO. The estimators discussed in this paper build on these results and lead to further improvements.
While we focus on the IWAE objective, our proposed \gdregs estimator applies generally.

\paragraph{Gradient estimation.} 
In practice, the expectation in \cref{eq:objective} and its gradients are intractable, so we approximate them using Monte Carlo sampling,
which makes the estimates of the objective and its gradients random variables. The resulting gradient estimators will be unbiased but have non-zero variance. We prefer estimators with lower variance, as they enable fast training by allowing higher learning rates.

We can distinguish between two general types of gradient estimators in this setting: (i) \emph{score function estimators} and (ii) \emph{pathwise estimators}. \!\!\tikzunderline{decorate, decoration=snake, line width=1.5pt, C4}{Score functions} are gradients of a log probability density \wrt its parameters, such as $\grad{\vphi}\lqz$; they treat the function $f_{\vphi, \vtheta}(\vz)$ as a black box and often yield high variance gradients. In contrast, \!\!\tikzunderline{line width=1.5pt, C3}{pathwise} estimators move the parameter-dependence from the probability density into the argument $\vz$ of the function $f_{\vphi, \vtheta}(\vz)$ and derive the computation path to often achieve lower variance gradients by using the knowledge of $\grad{\vz}f_{\vphi, \vtheta}(\vz)$; see \citet{Mohamed2020_gradient_estimation_ml} for a recent review.

When computing gradients of the objective $\varL_{\vphi, \vtheta}$ we have to differentiate both the sampling distribution of the expectation, $\distqz$, as well as the function $f_{\vphi, \vtheta}(\vz)$, 
\begin{align}
    \begin{split}
    & \gradtd{\vphi}{\expect{\distqz}{f_{\vphi,\vtheta}(\vz)}}  = \\
    & \quad = \expectsym{\distqz}\big[\tikzunderline{decorate, decoration=snake, line width=1.5pt, C4}{$\grad{\vphi}{f_{\vphi,\vtheta}(\vz)}$} + f_{\vphi,\vtheta}(\vz) \tikzunderline{decorate, decoration=snake, line width=1.5pt, C4}{$\grad{\vphi}{\log\distqz}$} \big]
    \end{split}\label{eq:proposal_tdgradient}\\
    & \gradtd{\vtheta}{\expect{\distqz}{f_{\vphi,\vtheta}(\vz)}}  = \expect{\distqz}{\tikzunderline{decorate, decoration=snake, line width=1.5pt, C4}{$\grad{\vtheta}f_{\vphi,\vtheta}(\vz)$}}, \label{eq:prior_tdgradient}
\end{align}
and both can give rise to\tikzunderline{decorate, decoration=snake, line width=1.5pt, C4}{score functions}. To see that all of the underlined terms indeed contain score functions, note that we can rewrite $\grad{\vphi}{f_{\vphi,\vtheta}(\vz)}$ as $\grad{\vphi}{f_{\vphi,\vtheta}(\vz)} = \grad{\lqz}{f_{\vphi,\vtheta}(\vz)}\tikzunderline{decorate, decoration=snake, line width=1.5pt, C4}{$\grad{\vphi}{\log\distqz}$}$ and similarly for $\grad{\vtheta}{f_{\vphi,\vtheta}(\vz)} =\grad{\lpz}{f_{\vphi,\vtheta}(\vz)}\tikzunderline{decorate, decoration=snake, line width=1.5pt, C4}{$\grad{\vtheta}{\log\distpz}$}$.

In the following we recapitulate how to address the score functions \wrt $\vphi$ in \cref{eq:proposal_tdgradient} using the reparameterization trick  and doubly-reparameterized gradients (\dregs, \citet{Tucker2019_dregs}), respectively. In \cref{sec:gdregs} we introduce \gdregs, a generalization of \dregs, that eliminates the score function \wrt $\vtheta$ in \cref{eq:prior_tdgradient}.

\paragraph{Reparameterization.} We can use the \emph{reparameterization trick} \citep{Kingma_vae,Rezende2014_vae} to turn the score function $\grad{\vphi}{\log \distqz}$ inside the expectation %
in \cref{eq:proposal_tdgradient} into a pathwise derivative of the function $f_{\vphi, \vtheta}(\vz)$ as follows: we express the latent variables $\vz \sim \distqz$ through a bijection of new random variables $\veps \sim \distqeps$, which are independent of $\vphi$, 
\begin{align}
    \vz = \zqreparam \Leftrightarrow \veps = \mathcal{T}^{-1}_q(\vz; \vphi).
\end{align} 
This allows us to rewrite expectations \wrt $\distqz$ as $\expect{\distqz}{f_{\vphi,\vtheta}(\vz)} = \expect{\distqeps}{f_{\vphi,\vtheta}(\zqreparam)}$, which moves the parameter dependence into the argument of $f_{\vphi, \vtheta}(\vz)$ and gives rise to a \!\!\tikzunderline{line width=1.5pt, C3}{pathwise gradient}:
\begin{align}
    \begin{split}
    & \gradtd{\vphi}{\expect{\distqz}{f_{\vphi,\vtheta}(\vz)}}  = %
    \expectsym{\distqeps}\big[\tikzunderline{thick, decorate, decoration=snake, line width=1.5pt, C4}{$\grad{\vphi}{f_{\vphi,\vtheta}(\vz)}$}\big. + {}\\
    & \qquad \big.{} + \tikzunderline{thick, line width=1.5pt, C3}{$\grad{\vz}{f_{\vphi,\vtheta}(\vz)}\grad{\vphi}{\zqreparam}$}\big]_{\vz=\zqreparam} .
    \end{split}\label{eq:proposal_tdgradient_reparam}
\end{align}
In \cref{sec:dregs_hierarchical} we discuss that this seemingly pathwise gradient in \cref{eq:proposal_tdgradient_reparam} can actually contain score functions for more structured or hierarchical models and explain how to extend \dregs to this case. For the remainder of this section we restrict ourselves to simple (single stochastic layer) models.

\paragraph{Double reparameterization.} \citet{Tucker2019_dregs} further reduce gradient variance by replacing the remaining score function  in the reparameterized gradient \cref{eq:proposal_tdgradient_reparam}, \tikzunderline{thick, decorate, decoration=snake, line width=1.5pt, C4}{$\grad{\vphi}{f_{\vphi,\vtheta}(\vz)}$} ${}= \grad{\lqz}{f_{\vphi,\vtheta}(\vz)}\tikzunderline{decorate, decoration=snake, line width=1.5pt, C4}{$\grad{\vphi}{\log\distqz}$}$, with its reparameterized counterpart.
Double reparameterization is based on the identity \cref{eq:dregs_eps} (Eq. 5 in \citet{Tucker2019_dregs}),
\begin{align}
    &\hspace{-0.75em} \expect{\vz\sim\distqz}{g_{\vphi,\vtheta}(\vz)\scoreqz}  = \nonumber\\
    & \hspace{0.4em} = \expect{\veps\sim\distqeps}{\left.\gradtd{\vz}{g_{\vphi,\vtheta}(\vz)}\right\vert_{\vz=\zqreparam} \grad{\vphi}{\zqreparam}} \label{eq:dregs_eps}\\
    & \hspace{0.4em} = \expectsym{\vz\sim\distqz}\!\!\left[\gradtd{\vz}{g_{\vphi,\vtheta}(\vz)} \big.\!\grad{\vphi}{\reparamq}\big\vert_{\veps = \mathcal{T}^{-1}_q(\vz; \vphi)}\right] \label{eq:dregs_z} \raisetag{1.5em}
\end{align}
which follows from the fact that both the score function and the reparameterization estimators are unbiased and thus equal in expectation. %
This identity holds for arbitrary $g_{\vphi,\vtheta}(\vz)$; to match the score function in \cref{eq:proposal_tdgradient_reparam} with the LHS of \cref{eq:dregs_eps}, we have to choose  $g_{\vphi,\vtheta}(\vz)=\grad{\lqz}{f_{\vphi,\vtheta}(\vz)}$. 

In \cref{eq:dregs_z} we have rewritten the expectation over $\veps\sim\distqeps$ in terms of $\vz\sim\distqz$, as this will become useful for our later generalization. Note how, to compute the pathwise gradient, the sample $\vz$ is mapped back to the noise variable $\veps = \reparamqinv$. $\grad{\vphi}{\mathcal{T}_q\left(\veps, \vphi\right)}$ is also sometimes written as $\grad{\vphi}{\vz(\veps; \vphi)}$ \citep{Tucker2019_dregs}.

\paragraph{Gradient estimation for the IWAE objective.} For the IWAE objective \cref{eq:iwae}, \citet{Tucker2019_dregs} derived the following doubly-reparametererized gradients (\dregs) estimator, which supersedes the previously proposed \stl estimator \citep{Roeder2017_stl}:
\begin{align}
    \hspace{-0.5em}\dregsgrad{\vphi}{\liwae} & =\textstyle\sum_{k=1}^K \normwk^2 \gradtd{\vz_k}{\log \wk} \grad{\vphi}{\mathcal{T}_q(\veps_k; \vphi)}
    \label{eq:IWAE_DReGs} \\
    \stlgrad{\vphi}{\liwae} & =\textstyle\sum_{k=1}^K \normwk \gradtd{\vz_k}{\log \wk} \grad{\vphi}{\mathcal{T}_q(\veps_k; \vphi)}
    \label{eq:iwae_stl}
\end{align}
with normalized importance weights $\normwk = \tfrac{\wk}{\sum_{k'=1}^K \wkp}$ and $\veps_{1:K}\sim \distqeps$. While the \dregs estimator doubly-reparameterizes the score function in \cref{eq:proposal_tdgradient_reparam}, the \stl estimator simply drops it and is biased as a result.
Crucially, because \dregs relies on reparameterization, it is limited to score functions of the sampling distribution $\distqz$, making it inapplicable in the more general setting of arbitrary score functions, such as $\expect{\distqz}{\grad{\vtheta}f_{\vphi,\vtheta}(\vz)}$ in \cref{eq:prior_tdgradient}. %

\section{\dregs for hierarchical models}
\label{sec:dregs_hierarchical}

We now show that for models with hierarchically structured latent variables even terms that look like pathwise gradients, such as $\gradtd{\vz}{f_{\vphi, \vtheta}(\vz)}$ in \cref{eq:proposal_tdgradient_reparam} or $\gradtd{\vz_k}{\log\wk}$ in the \dregs or \stl estimator for the IWAE objective \cref{eq:IWAE_DReGs,eq:iwae_stl}, can give rise to additional score functions. %
\emph{These additional score functions appear because the distribution parameters of 
one stochastic layer depend on the latent variables of another layer}. Their appearance is contrary to the intuition that doubly-reparameterized gradient estimators only contain pathwise gradients.

\subsection{An illustrative example}

To illustrate this, consider a hierarchical model with two layers where we first sample $\vz_2 \sim q_{\vphi_2}(\vz_2)$ and then $\vz_1 \sim q_{\vphi_1}(\vz_1 \given \vz_2)$.%
\footnote{The subscript indices refer to the latent layer indices and not to the importance samples in this case.} Note that the conditioning on $\vz_2$ is through the distribution parameters of $q_{\vphi_1}(\vz_1 \given \vz_2)$; to highlight this dependence of $\vz_1$ on $\vz_2$, we rewrite $q_{\vphi_1}(\vz_1 \given \vz_2) = q_{\valpha_{1\given 2}(\vz_2, \vphi_1)}(\vz_1)$, where we explicitly distinguish between the \emph{distribution} parameters $\valpha_{1\given 2}$, such as the mean and covariance of a Gaussian, and the \emph{network} parameters $\vphi_1$ that parameterize them together with the previously sampled latent $\vz_2$. %
A derivative \wrt $\vz_2$ that looks like a pathwise gradient actually gives rise to a~\!\!\tikzunderline{thick, dashed, line width=1.5pt, C4}{score function term} at a subsequent~level:
\begin{align}
    & \gradtd{\vz_2}{\log q_{\vphi_1}(\vz_1 \given \vz_2) } = \gradtd{\vz_2}{\log q_{\valpha_{1\given 2}(\vz_2, \vphi_1)}(\vz_1)} \\
    & \quad = \tikzunderline{thick, dashed, line width=1.5pt, C4}{$\grad{\valpha_{1\given 2}}{\log q_{\valpha_{1\given 2}}(\vz_1)}$}\grad{\vz_2}{\valpha_{1\given 2}(\vz_2, \vphi_1)} + \dots . \label{eq:hierarchical_offending_gradients}
\end{align}
We omitted (true) pathwise gradients ($\dots$), as the samples $\vz_1$ also depend on $\vz_2$ through reparameterization.
Similar additional score functions arise for seemingly pathwise gradients of hierarchical or autoregressive priors and variational posteriors.

\subsection{Extending \dregs to hierarchical VAEs}

Here we show how to extend \dregs to hierarchical VAEs to effectively reduce gradient variance for the variational posterior despite the results in the previous section. We still consider the IWAE objective (\cref{eq:iwae}), but now the latent space $\vz$ is structured, and both $p_{\vtheta}$ and $q_{\vphi}$ are hierarchically factorized distributions.

Let us consider a $2$-layer VAE 
\tikz[baseline=($(z2.south) ! 0.75 ! (z2.base)$)]{
    \node[shape=circle, draw, inner sep=0.5pt] (z2) {$\vz_2$};
	\node[shape=circle, draw,right=0.5of z2, inner sep=0.5pt] (z1) {$\vz_1$};
	\node[shape=circle, draw, right=0.5of z1, inner sep=1.7pt] (x) {$\vx$};
	\draw[-stealth', line width=.75pt] (z2) -- (z1);
	\draw[-stealth', line width=.75pt] (z1) -- (x);
	\useasboundingbox (z2.south west) rectangle (x.base east);
	}
and examine the term $\gradtd{\vphi_2}{\log q_{\vphi_1, \vphi_2}(\vz_1, \vz_2)}$, which appears in the total derivative of the IWAE objective, %
as a concrete example. 
We have sampled $\vz_1$ and $\vz_2$ hierarchically using reparameterization:
$\vz_2(\vphi_2) \equiv \mathcal{T}_{q_2}\left(\veps_2; \valpha_2(\vphi_2)\right)$ and
$\vz_1(\vphi_1, \vphi_2) \equiv \mathcal{T}_{q_1}(\veps_1; \valpha_{1\given 2}(\vz_2(\vphi_2), \vphi_1))$:
\begin{align}
    \gradtd{\vphi_2}{\log\! \left[q_{\tikzunderline{thick, decorate, decoration=snake, line width=1.5pt, C4}{\small$\valpha_2(\vphi_2)$}}\!\!\left(\tikzunderline{thick, line width=1.5pt, C3}{$\vz_2(\vphi_2)$}\right) q_{\tikzunderline{thick, dashed, line width=1.5pt, C4}{\small$\valpha_{1 \given 2}(\vz_2(\vphi_2), \vphi_1)$}}\!\!\left(\tikzunderline{thick, line width=1.5pt, C3}{$\vz_1(\vphi_1, \vphi_2)$}\right)\right]} \label{eq:pathwise_actually_score}\raisetag{3.em}
\end{align}
\begin{figure*}[!b]
\begin{tcolorbox}[enhanced,colback=white,%
    colframe=red!75!black, attach boxed title to top right={yshift=-\tcboxedtitleheight/2, xshift=-1.25cm}, title=\gdregs identity, coltitle=red!75!black, boxed title style={size=small,colback=white,opacityback=1, opacityframe=0}, size=title, enlarge top initially by=-\tcboxedtitleheight/2]
 \vskip-0.75em
\begin{equation}
    \hspace*{-.35cm}\expectsym{\vz\sim\distqz}\!\left[\gz\scorepz\right] = \expectsym{\vz\sim\distqz}\!\left[\!\left(\gz\gradtd{\vz}{\log \tfrac{\distqz}{\distpz}} + \gradtd{\vz}{\gz}\right)\left.\!\!\grad{\vtheta}{\mathcal{T}_p(\vepst; \vtheta)}\right\vert_{\vepst = \mathcal{T}_p^{-1}(\vz, \vtheta)}\right]
\label{eq:GDReGs}
\end{equation}
\end{tcolorbox}
\vskip-1.75em
\begin{align}
    & \gradtd{\vtheta}{\expect{\distqz}{\gz}}  \aboveeq{\circled{inner sep=1pt}{\tiny \textsf{1}}} \gradtd{\vtheta}{\expect{\distpz}{\tfrac{\distqz}{\distpz}\gz}} 
    \aboveeq{\circled{inner sep=1pt}{\tiny \textsf{1}}} \gradtd{\vtheta}{\expect{\distqepst}{\tfrac{\distq(\reparampt)}{\distp(\reparampt)} g_{\vphi, \vtheta}\!\left(\reparampt\right)}}; \quad \distqepst = \mathcal{N}(0, \mathbb{I}) \\
    & \quad \aboveeq{\circled{inner sep=1pt}{\tiny \textsf{2}}} \expectsym{\distqepst}\Big[   {\gradtd{\vz}{\left(\tfrac{\distqz}{\distpz}g(\vz)\right)}\grad{\vtheta}\reparampt} + \tfrac{\distqz}{\distpz}\big(\grad{\vtheta}{\gz}-g(\vz)\scorepz\big) \Big]_{\vz=\reparampt}\\
    & \quad \aboveeq{\circled{inner sep=1pt}{\tiny \textsf{3}}} \expectsym{\distqz}\Big[ \left(g(\vz)\gradtd{\vz}{\log\tfrac{\distqz}{\distpz}}+\gradtd{\vz}{g(\vz)}\right)\left.\grad{\vtheta}\reparampt\right\vert_{\vepst = \reparampinv} + \grad{\vtheta}{\gz} - g(\vz)\scorepz \Big]
\end{align}
\vskip-1.25em
\caption[]{The \gdregs identity and a brief derivation in three steps: \protect\circled{inner sep=1pt}{\tiny \textsf{1}} \emph{temporarily} change the path so that it depends on $\vtheta$; \protect\circled{inner sep=1pt}{\tiny \textsf{2}} perform the reparameterized gradient computation; \protect\circled{inner sep=1pt}{\tiny \textsf{3}} change the path back so we can use samples $\vz\sim\distqz$ to estimate the expectation. See \cref{app:sec:GDReGs} for details and an alternative derivation using the change of density formula.}
\label{fig:gdregs_identity}
\end{figure*}
When computing total derivatives \wrt parameters $\vphi_2$ of the upper layer, we distinguish between three types of gradients: the
(true) \!\!\tikzunderline{thick, line width=1.5pt, C3}{pathwise gradients \wrt $\vz_1$ and $\vz_2$}, a \!\!\uldscore{\emph{direct}} \uldscore{score function} because the distribution parameters $\valpha_2(\vphi_2)$ directly depend on $\vphi_2$, and an \!\!\ulidscore{\emph{indirect} score function} because the distribution parameter $\valpha_{1 \given 2}(\vz_2(\vphi_2), \vphi_1)$ indirectly depends on $\vphi_2$ through $\vz_2(\vphi_2)$. %
Indirect score functions do not arise in single stochastic layer models considered by \citet{Tucker2019_dregs}, and we have three options to estimate them:
(1) leave them---this naive estimator is unbiased but potentially has high variance; (2) drop them, similar to \stl{}---this estimator is generally biased; (3) doubly-reparameterize them using \dregs{} again---this estimator is unbiased, but can generate further indirect score functions.

Total derivatives of other terms in the objective similarly decompose into pathwise gradients as well as direct and indirect score functions. Notably, this includes indirect score functions of the prior $\log p_{\vtheta_1, \vtheta_2}(\vz_1, \vz_2)$, to which \dregs does not apply. In \cref{sec:gdregs} we introduce the generalized \dregs (\gdregs) estimator that applies in this case.

\subsection{\dregs for hierarchical IWAE objectives}

For IWAE objectives we find that the indirect score functions come up twice: once when computing pathwise gradients of the initial reparameterization, and a second time (with a different prefactor) when computing pathwise gradients for the double-reparameterization of the direct score functions. The same happens for the (true) pathwise gradients, and it is this double-appearance and the resulting cancellation of prefactors that helps reduce gradient variance for \dregs. Moreover, for general model structures it is impossible to replace all successively arising indirect score functions with pathwise gradients, even by applying \gdregs. For example, when the prior and posterior do not factorize in the same way, double-reparameterization of one continuously creates indirect score functions of the other and vice versa, see \cref{app:sec:hierarchical_dregs_gdregs,app:sec:double_reparameterization_indirect_score} for more details. 

Thus, to extend \dregs to hierarchical models, we leave the indirect score functions unchanged and only doubly reparameterize the direct score functions. 
The extended \dregs estimator for IWAE models with arbitrary hierarchical structures is given by \cref{eq:iwae_hierarchical_dregs} 
\begin{tcolorbox}[enhanced,colback=white,%
    colframe=C1!75!black, attach boxed title to top right={yshift=-\tcboxedtitleheight/2, xshift=-.5cm}, title=\dregs estimator for hierarchical IWAE, coltitle=C1!75!black, boxed title style={size=small,colback=white,opacityback=1, opacityframe=0}, size=title, enlarge top initially by=-\tcboxedtitleheight/2, left=-5pt]
 \vskip-1.em
\begin{align}
    & \dregsgrad{\vphi_l}{\liwae} = \qquad \qquad \qquad \veps_{\{1:K\}l}\sim\distqeps 
\label{eq:iwae_hierarchical_dregs} \\
    & \quad =\sum_{k=1}^K \normwk^2\gradtd{\zkl}{\log\wk}\grad{\vphi_l}{\reparam{q_l}{\epskl; \valpha_l(\pa{\valpha}{l}, \vphi_l)}} \nonumber
\end{align}
\end{tcolorbox}
where $l$ denotes the stochastic layer, $\pa{\valpha}{l}$ is the set of latent variables that $\zkl$ depends on, and $\vz_{kl} = \mathcal{T}_{q_l}(\veps_{kl}; \valpha_l(\pa{\valpha}{l}, \vphi_{l}))$ through reparameterization.
We provide a detailed derivation in \cref{app:sec:hierarchical_dregs_gdregs} and a worked example for a VAE with two stochastic layers in \cref{app:sec:worked_hierarchical_example}. In \cref{app:sec:objectives,app:sec:objectives_dregs} we show how to implement this estimator using automatic differentiation by using a \emph{surrogate loss function}, whose forward computation we discard, but whose backward computation exactly corresponds to the estimator in \cref{eq:iwae_hierarchical_dregs}. Alternatively, one could implement a custom gradient for the objective that directly implements \cref{eq:iwae_hierarchical_dregs}; however, we found our approach using a surrogate loss function to be simpler both conceptually and implementation-wise.

\citet{Roeder2017_stl} apply the \stl estimator to hierarchical ELBO objectives but do not discuss indirect score functions. Their experimental results are consistent with maintaining the indirect score functions, similar to how we extend \dregs to hierarchical models; the \stl estimator is biased for IWAE objectives \citep{Tucker2019_dregs}.%

\begin{figure*}[htb]
    \centering
    \includestandalone[mode=buildnew]{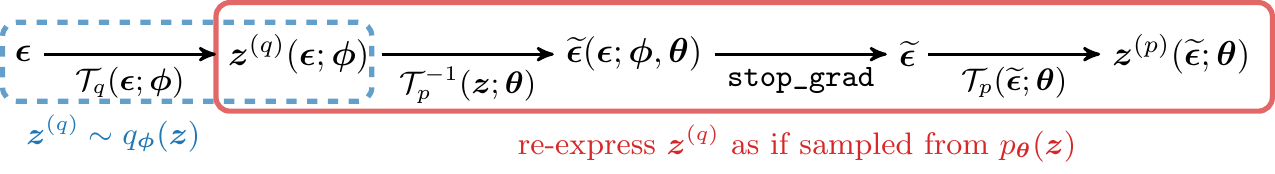}
    \vskip-0.75em
    \caption{Computational flow to re-express a sample $\vz$ from $q_{\vphi}(\vz)$ as if it were sampled from $p_{\vtheta}(\vz)$. Its numerical value and distribution remain unchanged but the pathwise gradient through it now depends on $\vtheta$: $\left.\grad{\vtheta}{\mathcal{T}_p(\vepst; \vtheta)}\right\vert_{\vepst = \mathcal{T}_p^{-1}(\vz, \vtheta)}$.
    $\vepst= \mathcal{T}_p^{-1}(\mathcal{T}_q(\veps; \vphi); \vtheta)$ follows a different, usually more complex, distribution from $\veps\sim\distqeps$.%
    }
    \label{fig:resampled_as_if_from_schematic_single}
\end{figure*}

\section{Generalized \dregs}
\label{sec:gdregs}
Here, we generalize \dregs to score functions that involve distributions $\distpz$ different from the sampling distribution $\distqz$, as in \cref{eq:prior_tdgradient}. In other words, we would like to replace score function terms of the form $\expect{\distqz}{g_{\vphi,\vtheta}(\vz)\scorepz}$ with doubly-reparameterized pathwise gradients. Such terms appear, for example, when training a VAE with a trainable prior $\distpz$ with the ELBO or IWAE objectives. %
\dregs cannot be used directly in this case as it relies on reparameterization of the sampling distribution $\distqz$, so that the path depends on the parameters $\vphi$, whereas the score function is with \wrt parameters $\vtheta$ of a different distribution $\distpz$. 

To make progress \emph{we need to make the path depend on the parameters $\vtheta$} while still sampling $\vz\sim\distqz$ during training. Our solution consists of three steps (also see \cref{fig:gdregs_identity}): 
\begin{itemize}[topsep=-2pt,itemsep=2pt,partopsep=0pt, parsep=0pt, leftmargin=2em]
\item[\circled{inner sep=1pt}{\tiny \textsf{1}}] \emph{temporarily} change the path so that it depends on $\vtheta$; 
\item[\circled{inner sep=1pt}{\tiny \textsf{2}}] perform the reparameterized gradient computation; 
\item[\circled{inner sep=1pt}{\tiny \textsf{3}}] change the path back so we can use samples $\vz\sim\distqz$ to estimate the expectation. 
\end{itemize}
We change the path by first using an importance sampling reweighting to temporarily re-write the expectation, $\expect{\distqz}{\ast} = \expect{\distpz}{\frac{\distqz}{\distpz} \ast}$, and then employing reparameterization on the new sampling distribution $\distpz$: $\vz = \reparampt$ with $\vepst\sim\distqepst$. Following this recipe, we derive the gradient identity in \cref{eq:GDReGs} for general $g_{\vphi,\vtheta}(\vz)$ that we refer to as generalized \dregs (or \gdregs for short) identity.

Like \dregs (\cref{eq:dregs_z}), \gdregs allows us to transform score functions into pathwise gradients. Yet, unlike \dregs, \gdregs applies to general score functions and contains a correction term that vanishes when $\distpz$ and $\distqz$ are identical ($\log\tfrac{\distqz}{\distpz}$ term in \cref{eq:GDReGs}).

\begin{figure*}[htb]
    \centering
    \includestandalone[mode=buildnew]{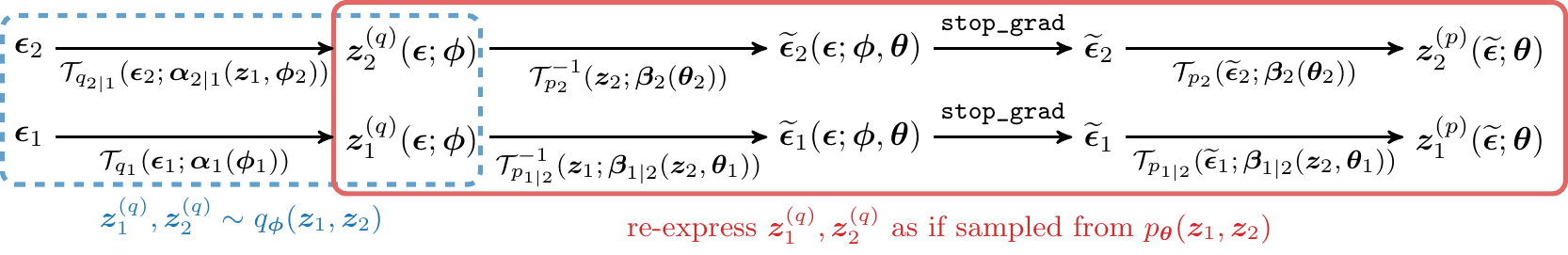}
    \vskip-0.75em
    \caption{Computational flow to re-express samples $\vz_1, \vz_2$ from $q_{\vphi}(\vz_1, \vz_2) = q_{\vphi_1}(\vz_1)q_{\vphi_2}(\vz_2 \given \vz_1)$ as if they were sampled from $p_{\vtheta}(\vz_1, \vz_2) = p_{\vtheta_2}(\vz_2)p_{\vtheta_1}(\vz_1\given \vz_2)$. Their numerical values and distribution remain unchanged but the gradient flow through them changes. Note that ${\vepst_i}$ follows a different, usually more complex, distribution  from $\veps_i$. $\valpha_i$ and $\vbeta_i$ denote the distribution parameters of the variatonal posterior and the prior, respectively.}
    \label{fig:resampled_as_if_from_schematic}
\end{figure*}

Note that the pathwise derivative $\grad{\vtheta}{\reparampt}$ in \cref{eq:GDReGs} looks like we reparameterized an independent noise variable $\vepst$ using $\distpz$, where the numerical value of the noise variable is given by $\vepst=\reparampinv$ and $\vz\sim\distqz$. We can interpret this sequence of transformations as a normalizing flow \citep{Rezende2015_flows} $\vz\rightarrow\vepst\rightarrow\vz$, 
such that $\mathcal{T}_p(\vepst; \vtheta) = \mathcal{T}_p(\mathcal{T}_p^{-1}(\vz; \vtheta); \vtheta) = \vz$. We can think of this procedure as \emph{re-expressing the sample $\vz \sim \distqz$ as if it came from $\distpz$}: Its numerical value $\vz$ remains unchanged and it is still distributed according to $\distqz$, yet its pathwise gradient $\grad{\vtheta}{\reparampt}$ depends on $\vtheta$. We illustrate the corresponding computational flow in \cref{fig:resampled_as_if_from_schematic_single} and provide an example implementation with code in \cref{app:sec:code}

Note that to derive the \gdregs identity, we only require $\distpz$ to be reparameterizable 
(\tikz[baseline=(a.base)]{\node (a) {}; \draw[line width=1.5pt, C4light, rounded corners=0.5mm] (0,0) rectangle ++(2em,.85em) node[pos=.1](a){};} in \cref{fig:resampled_as_if_from_schematic_single}). 
While $\distqz$ may be reparameterizable as well (\tikz[baseline=(a.base)]{\node (a) {}; \draw[line width=1.5pt, C1light, rounded corners=0.5mm, dashed] (0,0) rectangle ++(2em,.85em) node[pos=.1](a){};} in \cref{fig:resampled_as_if_from_schematic_single}), 
this is not necessary; we only need to be able to evaluate its density in \cref{eq:GDReGs}.
In the simplest case of the cross-entropy objective $\mathcal{L}^\text{ce}=\expect{\distqz}{\log\distpz}$ (as in the ELBO with a sample-based KL estimate), $g_{\vphi, \vtheta}(\vz) = 1$, and the \gdregs identity \cref{eq:GDReGs} gives rise to the following \gdregs estimator:
\begin{align}
\hspace*{-0.4em}
    \gdregsgrad{\vtheta}{\mathcal{L}^\text{ce}} = \grad{\vz}{\log \tfrac{\distqz}{\distpz}}\Big.\grad{\vtheta}{\mathcal{T}_p(\vepst; \vtheta)}\Big\vert_{\vepst = \mathcal{T}_p^{-1}(\vz, \vtheta)}
\label{eq:gdregs:x-entropy}\raisetag{2.5em}
\end{align}
with $\vz\sim\distqz$. For Gaussian distributions $\distqz$ and $\distpz$, the cross-entropy and its gradients can be computed in closed form, which we can think of as a perfect estimator with zero bias and zero variance. Moreover, the expectation and variance of both the naive score function as well as the \gdregs estimator in \cref{eq:gdregs:x-entropy} can be computed in closed form. We provide full derivations and a discussion of this special case in \cref{app:sec:x-entropy} as well as an example implementation in terms of (pseudo-)code in \cref{app:sec:code}. The main results are: (i) \gdregs has lower variance gradients than the score function when $\distqz$ and $\distpz$ overlap substantially, %
which is typically the case at the beginning of training; (ii) we can derive a closed-form control variate that depends on a ratio of the means and variances of the two distributions and that is strictly superior to the naive score function estimator and the \gdregs estimator in terms of gradient variance. However, the analytic expression for the cross-entropy has even lower (zero) gradient variance in this case.

\subsection{\gdregs for VAE objectives}
\label{sec:gdregs_iwae}

We can now use the \gdregs identity \cref{eq:GDReGs} to derive generalized doubly-reparameterized estimators for expectations of general score functions of the form $\expect{\distqz}{g_{\vphi,\vtheta}(\vz)\scorepz}$, %
also see \cref{eq:prior_tdgradient}. In \cref{app:sec:gdregs_iwae} we derive the following \gdregs estimator of the IWAE objective \wrt the prior parameters $\vtheta$:
\begin{tcolorbox}[enhanced,colback=white,%
    colframe=C1!75!black, attach boxed title to top right={yshift=-\tcboxedtitleheight/2, xshift=-.5cm}, title=\gdregs estimator for IWAE, coltitle=C1!75!black, boxed title style={size=small,colback=white,opacityback=1, opacityframe=0}, size=title, enlarge top initially by=-\tcboxedtitleheight/2, left=-5pt]
 \vskip-1.em
\begin{align}
\begin{split}
& \hspace*{-0.2em} \gdregsgrad{\vtheta}{\liwae}  =  \textstyle\sum_{k=1}^K \big(\normwk \gradtd{\vz_k}\log p_{\vlambda}(\vx \given \vz_k) - \big.\\
   &\qquad -\Big.\big. \normwk^2 \gradtd{\vz_k}{\log w_k}\big) \grad{\vtheta}{\mathcal{T}_p(\vepst_k; \vtheta)}\Big\vert_{\vepst_k = \mathcal{T}_p^{-1}(\vz_k, \vtheta)}
\end{split}
\label{eq:IWAE_GDReGs}\raisetag{2.5em}  %
\end{align}
\end{tcolorbox}
with $\vz_{1:K}\sim\qzx$. The second term in \cref{eq:IWAE_GDReGs} looks like the \dregs estimator for $\vphi$ in \cref{eq:IWAE_DReGs} except that the samples $\vz_k$ are now re-expressed as if they came from $\distpz$. In addition we obtain a term that involves the likelihood $\likelzl$ and is linear in $\normwk$. 
Note that we do not apply \gdregs to the likelihood parameters $\vlambda$ because $\likelzl$ is a distribution over $\vx$ rather than $\vz$; in the following we therefore drop the subscript $\vlambda$. %

We learn all parameters by optimizing the same objective function \cref{eq:iwae}, but employ different gradient estimators for different subsets of parameters. In practice, we implement these estimators using different \emph{surrogate objectives} for the likelihood, proposal, and prior parameters, see \cref{app:sec:objectives} for details.
While separate objectives seem computationally expensive, most terms are shared between them, and modern frameworks avoid such duplicate computation. In practice, we found the runtime increase for training with \dregs and \gdregs estimators to be smaller than $10\%$ without any optimization of the implementation.

\subsection{Extending \gdregs to hierarchical VAEs}
When extending \gdregs to hierarchical models, we again encounter direct and indirect score functions (see \cref{sec:dregs_hierarchical}). Like for the posterior parameter $\vphi$ we apply \gdregs to the direct score functions but leave the indirect score functions unchanged. The full \gdregs estimator for IWAE objectives with arbitrary hierarchial structure is given in \cref{app:sec:hierarchical_dregs_gdregs} \cref{app:eq:iwae_hierarchical_gdregs}, see \cref{app:sec:hierarchical_dregs_gdregs} for a derivation. In \cref{app:sec:worked_hierarchical_example} we provide a worked example and in \cref{app:sec:objectives} we again show how to use surrogate losses to implement the estimator in practice.%

To apply \gdregs we need to re-express samples from $\distqz$ as if they came from $\distpz$. We do this for the entire hierarchy jointly. %
In \cref{fig:resampled_as_if_from_schematic} we illustrate the necessary computational flow for the example of a $2$-layer VAE with the variational posterior factorized in the opposite direction from the generative process; see \cref{app:sec:hierarchical_dregs_gdregs} for the general case. We draw samples $\vz_1, \vz_2 \sim q_{\vphi}(\vz_1, \vz_2) = q_{\vphi_1}(\vz_1)q_{\vphi_2}(\vz_2 \given \vz_1)$ (by transforming independent variables ${\veps_i}$) and then re-express them as if they were sampled from the prior $p_{\vtheta_2}(\vz_2)p_{\vtheta_1}(\vz_1\given \vz_2)$, which factorizes in the opposite direction. While the numerical values of $\vz_1$ and $\vz_2$ remain unchanged, $\vz_1$ is now dependent on $\vz_2$ and both depend on the respective $\vtheta$ parameters when computing gradients; we can view $(\vz_1, \vz_2)$ as samples that were obtained by transforming independent variables $(\vepst_1, \vepst_2)$ that follow a more complicated distribution than $(\veps_1, \veps_2)$. As in the single-layer case, only $\distpz$ needs to be reparameterizable.

\section{Experiments}

In this section we empirically evaluate the hierarchical extension of \dregs and its generalization to \gdregs, and compare them to the naive IWAE gradient estimator (labelled as \iwae) as well as \stl \citep{Roeder2017_stl}. First, we illustrate that \dregs and \gdregs increase the gradient signal-to-noise ratio (SNR) and reduce gradient variance compared to the naive estimator on a simple hierarchical example (\cref{sec:exp:toy}); second, we show that they also reduce gradient variance in practice and improve test performance on several generative modelling tasks with VAEs with one or more stochastic layers (\cref{sec:exp:generativemodelling}). We highlight that both the extension of \dregs to more than one stochastic layer as well as training the prior with \gdregs  are novel contributions of this work.

\subsection{Illustrative example: linear VAE}
\label{sec:exp:toy}
We first consider an extended version of the illustrative example by \citet{Rainforth2018_bound} and \citet{Tucker2019_dregs} to show that hierarchical \dregs and \gdregs increase the gradient signal-to-noise ratio (SNR) and reduce gradient variance compared to the naive IWAE gradient estimator.

We consider a $2$-layer linear VAE with hierarchical prior $\vz_2 \sim \mathcal{N}(0, \mathbb{I})$, $\vz_1 \given \vz_2 \sim \mathcal{N}(\vmu_{\vtheta}(\vz_2), \vsigma^2_{\vtheta}(\vz_2))$, Gaussian noise likelihood $\vx\given\vz_1 \sim \mathcal{N}(\vz_1, \mathbb{I})$, and bottom up variational posterior $q_{\vphi_1}(\vz_1 \given \vx) = \mathcal{N}(\vmu_{\vphi_1}(\vx), \vsigma^2_{\vphi_1}(\vx))$ and $q_{\vphi_2}(\vz_2 \given \vz_1) = \mathcal{N}(\vmu_{\vphi_2}(\vz_1), \vsigma^2_{\vphi_2}(\vz_1))$. All $\vmu_i$ and $\vsigma_i$ are linear functions, and $\vz_1, \vz_2, \vx \in\mathbb{R}^D$. We sample $512$ datapoints in $D=5$ dimensions from a model with $\vmu_{\vtheta}(\vz_2)=\vz_2$ and $\vsigma_{\vtheta}(\vz_2) = 1$. We then train the parameters $\vphi$ and $\vtheta$ using SGD and the IWAE objective til convergence and evaluate the gradient variance and signal-to-noise ratio for each estimator. For the proposal parameters $\vphi$ we compare \dregs to the naive score function (labelled as \iwae) and to \stl ; for the prior parameters $\vtheta$ we compare \gdregs to \iwae. 

\begin{figure}[tb]
    \centering
    \includestandalone[mode=buildnew]{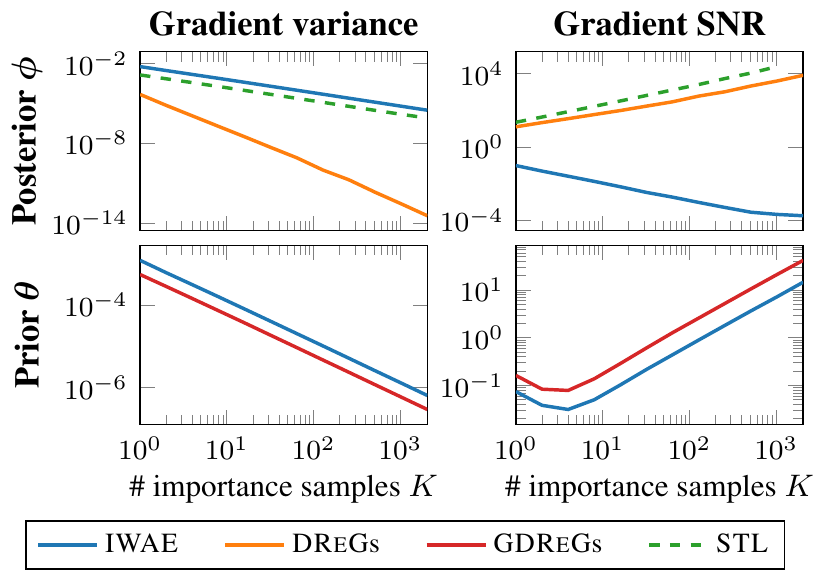}
    \vskip-.5em
    \caption{Average gradient variance \textit{(left)} and signal-to-noise ratio (SNR) \textit{(right)} for the proposal parameters $\vphi$ \textit{(top)} and the prior parameters $\vtheta$ \textit{(bottom)}.}
    \label{fig:toy_2layer}
\end{figure}

We find that our extension of \dregs to hierarchical models behaves qualitatively 
the same as in the single layer case considered by \citet{Tucker2019_dregs}, see \cref{fig:toy_2layer} \textit{(top)}: While the SNR for the naive estimator %
vanishes with increasing number of importance samples, the SNR increases for \dregs. %
This can be explained by the faster rate with which the gradient variance decreases for \dregs compared to \iwae and \stl. While \stl has an even better SNR, %
its gradients are biased. %

When considering gradients \wrt the prior parameters we find that the SNR is higher and the gradient variance is lower for \gdregs compared to the naive estimator (\iwae), see \cref{fig:toy_2layer} \textit{(bottom)}. However, they grow and shrink at the same rate for both estimators as the number of importance samples is increased.

\subsection{Image modelling with VAEs}
\label{sec:exp:generativemodelling}
In the remainder of this paper we consider image modelling tasks with VAEs on several standard benchmark datasets: MNIST \citep{lecun2010mnist}, Omniglot \citep{Lake_Omniglot_2015}, and FashionMNIST \citep{FashionMNIST}. We use dynamically binarized versions of all datasets to minimize overfitting.

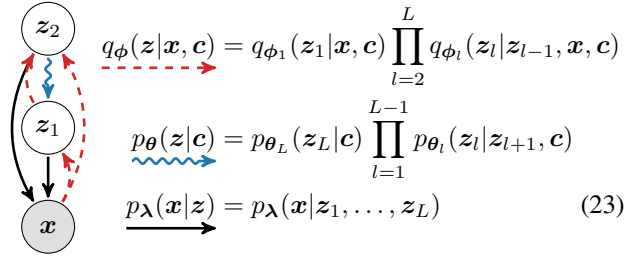
\begin{figure}[tb]
\begin{minipage}[t]{0.15\columnwidth}
\vskip-1.em
\begin{tikzpicture}[baseline={([yshift=1em]z2.north)}]
	\node[latent] (z2) {$\vz_2$};
	\node[latent, below = 0.6 of z2] (z1) {$\vz_1$};
	\node[obs, below = 0.6 of z1] (x) {$\vx$};
	\edge[decorate, decoration={snake, amplitude=0.1em, segment length=0.5em,post length=5pt, pre length=1pt}, C1]{z2}{z1};
	\edge{z1}{x};
	\edge[bend right]{z2}{x};
	\edge[bend left, dashed, C4]{z1}{z2};
	\edge[bend right, dashed, C4]{x}{z1};
	\edge[bend right, dashed, C4]{x}{z2};
	\useasboundingbox (x.south west) rectangle (z2.north east);
\end{tikzpicture}
\end{minipage}%
\begin{minipage}[t]{0.85\columnwidth}
\vskip-1.em
\begin{align}
    \tikzunderlinemoredist{dashed, line width=1pt, C4, -stealth'}{$\qzxc$} & = q_{\vphi_1}(\vz_1 \given \vx, \vc) \prod_{l=2}^L q_{\vphi_l}(\vz_l \given \vz_{l-1}, \vx, \vc) \nonumber\\
    \tikzunderlinemoredist{decorate, decoration={snake,post length=5pt}, line width=1pt, C1, -stealth'}{$\pzc$} & = p_{\vtheta_L}(\vz_L \given \vc) \prod_{l=1}^{L-1} p_{\vtheta_l}(\vz_l \given \vz_{l+1}, \vc) \nonumber\\
    \tikzunderlinemoredist{line width=1pt, -stealth'}{$\likelzl$} & = p_{\vlambda}(\vx \given \vz_1, \dots, \vz_L) \label{eq:vae_models}
\end{align}
\end{minipage}
\vskip-0.3em
\caption{Model specification and $2$-layer example for conditional and unconditional image modelling.}
\label{fig:vae_models}
\vskip-0.5em
\end{figure}

We consider both single layer and hierarchical (multi-layer) VAEs and evaluate them on unconditional and conditional modelling tasks using the IWAE objective, \cref{eq:iwae}. 
In the hierarchical case, the generative path (prior and likelihood) is top-down whereas the variational posterior is bottom-up, see \cref{fig:vae_models,eq:vae_models} for a full description of the model and a $2$-layer unconditional example.
For conditional modelling we predict the bottom half of an image given its top half, as in \citet{Tucker2019_dregs}; in this case, both the prior and variational posterior also depend on a context variable $\vc$, $\qzxc$ and $\pzc$, respectively. 
We use a factorized Bernoulli likelihood along with factorized Gaussians for the variational posterior and prior. Every conditional distribution in \cref{eq:vae_models} is parameterized by an MLP with two hidden layers of $300$ \texttt{tanh} units each, and all latent spaces have $50$ dimensions. Unless stated otherwise, we train all models for $1000$ epochs using the Adam optimizer \citep{Kingma_adam} with default learning rate of $3\cdot 10^{-4}$, a batch size of $64$, and $K=64$ importance samples; see \cref{app:sec:exp_details_results} for details.

\begin{figure*}[htb]
    \includestandalone[mode=buildnew]{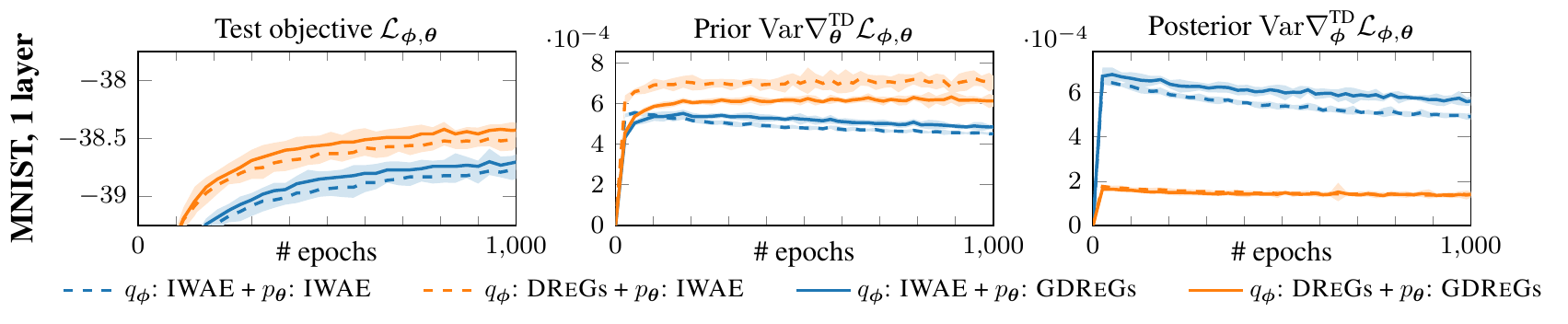} \\
    \includestandalone[mode=buildnew]{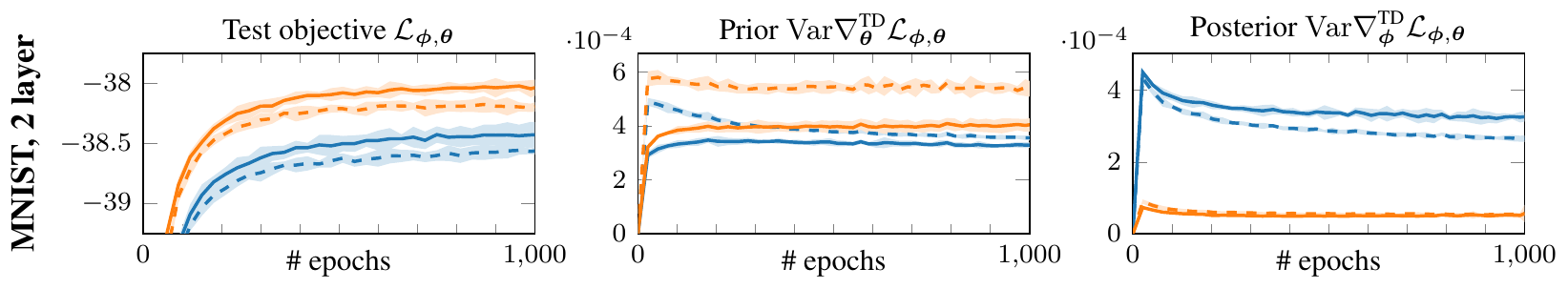}
    \vskip-0.5em
    \caption{\emph{Conditional} image modelling of MNIST with a VAE with $1$ layer (top) and $2$ layers (bottom). Shaded areas denote $\pm$ 1.96 standard deviations $\sigma$ over $5$ reruns.}
    \label{fig:exp:1_2layers_cond_mnist}
\end{figure*}

\begin{figure*}[htb]
    \centering
    \includestandalone[mode=buildnew]{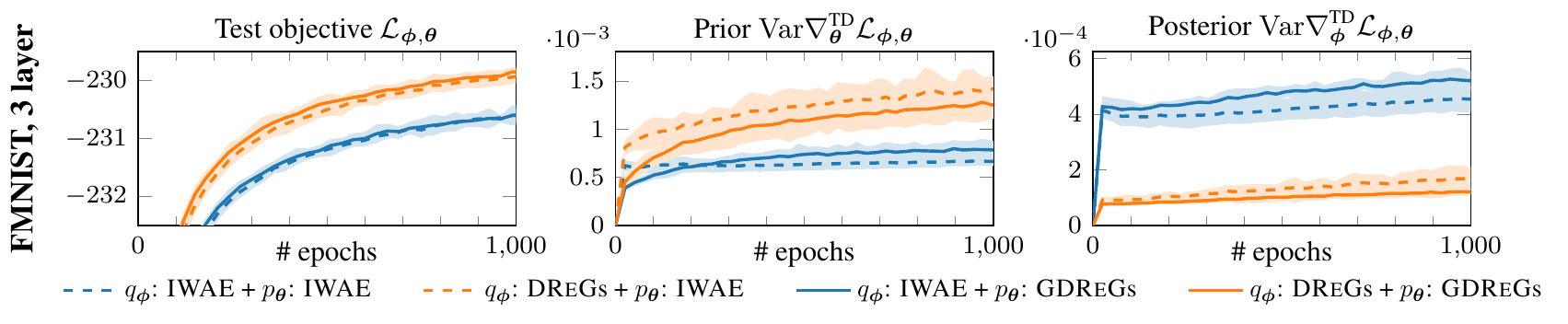}
    \vskip-0.5em
    \caption{\emph{Unconditional} image modelling on FashionMNIST; $3$ layers.}
    \label{fig:exp:3layers_fmnist}
\end{figure*}

As mentioned in \cref{sec:gdregs_iwae}, we use separate surrogate objectives to compute the gradient estimators for the likelihood, posterior, and prior parameters. While we always train the likelihood parameters $\vlambda$ on the naive IWAE objective, we consider the naive IWAE estimator (labelled as \iwae), \stl, and \dregs for the variational posterior parameters $\vphi$, and \iwae and \gdregs for the prior parameters $\vtheta$. See \cref{app:sec:objectives} for details on the implementation of the estimators.
We present the results for conditional modelling in \cref{tab:exp:alllayers_cond} and \cref{fig:exp:1_2layers_cond_mnist}, and for unconditional modelling in \cref{tab:exp:alllayers} and \cref{fig:exp:3layers_fmnist}; see \cref{app:sec:exp_details_results} for more experimental results. 

\paragraph{Estimators for the variational parameters $\vphi$.} First, we evaluate the choice of estimator for the parameters of \distqz. Like \citet{Tucker2019_dregs} for the single layer case, we find that our extension of \dregs to hierarchical models leads to a dramatic reduction in gradient variance for the variational posterior parameters $\vphi$ on all tasks (third column in \cref{fig:exp:1_2layers_cond_mnist,fig:exp:3layers_fmnist}), which translates to an improved test objective in all cases considered. \dregs is unbiased and typically outperforms the (biased) \stl estimator. We also observed similar improvements on the training objective.

\begin{table*}[htb]
    \centering
    \small
    \newcommand{\numlayer}[1]{$#1$ layer}
    \begin{tabular}{llCCCCCC}
    \toprule
        \multicolumn{2}{r}{\textbf{estimator} $\gradtd{\phi}{}$}& \multicolumn{2}{c}{\textbf{\iwae}} & \multicolumn{2}{c}{\textbf{\stl}} & \multicolumn{2}{c}{\textbf{\dregs}} \\
        \multicolumn{2}{r}{\textbf{estimator} $\gradtd{\theta}{}$}& \textbf{\iwae} & \textbf{\gdregs} & \textbf{\iwae} & \textbf{\iwae} & \textbf{\iwae} & \textbf{\gdregs} \\\midrule
        \textbf{MNIST}& \numlayer{1} & -38.77\pms{0.01} & -38.71\pms{0.02}& -38.76\pms{0.03} & -38.68\pms{0.03} & -38.50\pms{0.01} & \mathbf{-38.44\pms{0.01}}  \\
        & \numlayer{2} & -38.55\pms{0.02}&-38.42\pms{0.03}&-38.24\pms{0.02}&-38.14\pms{0.02}&-38.20\pms{0.01}& \mathbf{-38.02\pms{0.02}} \\
        & \numlayer{3} & -38.63\pms{0.01}&-38.44\pms{0.02}&-38.20\pms{0.01}&-38.10\pms{0.02}&-38.20\pms{0.01}& \mathbf{-38.04\pms{0.01}}\\\midrule
        \textbf{Omniglot} & \numlayer{1}  & -55.84\pms{0.02}& -55.66\pms{0.03}& -55.80\pms{0.05}&-55.62\pms{0.05}& -55.34\pms{0.02}&\mathbf{-55.24\pms{0.02}}\\
        & \numlayer{2} & -55.27\pms{0.03}&-54.98\pms{0.02}& -54.66\pms{0.03}& \mathbf{-54.28\pms{0.02}} &-54.73\pms{0.02}&-54.36\pms{0.03}\\
        & \numlayer{3}  & -55.35\pms{0.02}&-54.93\pms{0.02}&-54.64\pms{0.03}& \mathbf{-54.21\pms{0.03}} &-54.72\pms{0.02}&-54.28\pms{0.02} \\\midrule
        \textbf{FMNIST} & \numlayer{1}  & -102.84\pms{0.02}&-102.80\pms{0.02}& -102.99\pms{0.02} & -102.88\pms{0.02}&-102.61\pms{0.01}&\mathbf{-102.58\pms{0.01}}\\
        & \numlayer{2} &-102.74\pms{0.02}&-102.68\pms{0.01}& -102.65\pms{0.02} & -102.48\pms{0.03}&-102.40\pms{0.01}& \mathbf{-102.30\pms{0.02}}\\
        & \numlayer{3} & -102.86\pms{0.01} &-102.71\pms{0.01}& -102.68\pms{0.01} &-102.42\pms{0.02} & -102.46\pms{0.01}& \mathbf{-102.26\pms{0.01}}\\\bottomrule
    \end{tabular}
    \vskip-0.5em
    \caption{Test objective values (higher is better) on \emph{conditional} image modelling with a VAE model trained with IWAE. Higher is better; errorbars denote $\pm$ 1.96 standard errors ($\sigma/\sqrt{5}$) over $5$ reruns.}
    \label{tab:exp:alllayers_cond}
\end{table*}

\begin{table*}[htb]
    \centering
    \small
    \newcommand{\numlayer}[1]{$#1$ layer}
    \begin{tabular}{llCCCCCC}
    \toprule
        \multicolumn{2}{r}{\textbf{estimator} $\gradtd{\phi}{}$}& \multicolumn{2}{c}{\textbf{\iwae}} & \multicolumn{2}{c}{\textbf{\stl}} & \multicolumn{2}{c}{\textbf{\dregs}} \\
        \multicolumn{2}{r}{\textbf{estimator} $\gradtd{\theta}{}$}& \textbf{\iwae} & \textbf{\gdregs} & \textbf{\iwae} & \textbf{\gdregs} & \textbf{\iwae} & \textbf{\gdregs} \\\midrule
        \textbf{MNIST}  & \numlayer{2}& -86.07\pms{0.02}&-86.04\pms{0.03}&-85.29\pms{0.02}&\mathbf{-85.23\pms{0.03}} &\mathbf{-85.25\pms{0.02}}& -85.32\pms{0.02}\\
         & \numlayer{3} & -85.69\pms{0.02}&-85.70\pms{0.02}&-85.01\pms{0.03}&-84.94\pms{0.05}&\mathbf{-84.87\pms{0.03}}& \mathbf{-84.90\pms{0.04}}\\\midrule
        \textbf{Omniglot}  & \numlayer{2}&-105.20\pms{0.02} &-105.11\pms{0.02}&-104.10\pms{0.05}&\mathbf{-104.00\pms{0.05}}&-104.12\pms{0.05}&\mathbf{-104.05\pms{0.04}}\\
        & \numlayer{3} &-104.68\pms{0.02} &-104.71\pms{0.03}&-104.02\pms{0.02}&\mathbf{-103.55\pms{0.03}}&-104.71\pms{0.03}&\mathbf{-103.51\pms{0.06}}\\\midrule
        \textbf{FMNIST}  & \numlayer{2}& -230.65\pms{0.03} & -230.61\pms{0.02} & -230.14\pms{0.02} & \mathbf{-229.98\pms{0.02}} & -230.04\pms{0.03} & \mathbf{-229.98\pms{0.03}} \\
        & \numlayer{3}& -230.60\pms{0.03}&-230.59\pms{0.03}& -230.26\pms{0.04} &-229.92\pms{0.03} & -229.92\pms{0.02}& \mathbf{-229.87\pms{0.03}}\\\bottomrule
    \end{tabular}
    \vskip-0.5em
    \caption{Test objective values on \emph{unconditional} image modelling with a VAE model trained with IWAE.}
    \label{tab:exp:alllayers}
\end{table*}

\paragraph{Estimators for the prior parameters $\vtheta$.} Second, we consider the estimators for the $\vtheta$ parameters of the prior \distpz. Using the \gdregs estimator instead of the naive IWAE estimator consistently improves the train and test objective when combined with \emph{any} estimator for the variational posterior, especially for conditional image modelling with deeper models. For unconditional image modelling the improvements are only marginal, though using \gdregs never hurts. In terms of gradient variance for the prior parameters $\vtheta$, \gdregs consistently performs better in the beginning of training, when it always has lower variance. However, later in training this is only consistently true when also using the \dregs estimator for the variational posterior parameters $\vphi$. %
We hypothesize that the \gdregs estimator yields larger improvements for conditional modelling because the prior and posterior distribution are closer to each other due to the conditioning, and we saw that \gdregs works particularly well in this case for Gaussian distributions, also see \cref{app:sec:x-entropy}. To quantify this ``closeness'' we compared the KL of the variational posterior to the prior on the same dataset and found it to be about twice as large for unconditional modelling than for conditional modelling, see \cref{app:sec:exp_details_results}.

We also note that the gradient variance for the prior parameters $\vtheta$ is higher when using the \dregs estimator for the variational posterior parameters $\vphi$, compared to the naive IWAE estimator (compare orange and blue lines in the middle column of \cref{fig:exp:1_2layers_cond_mnist,fig:exp:3layers_fmnist}). This is an indirect effect of altered learning dynamics. We suspect that better posterior gradient estimates with \dregs lead to generative models that fit the data better, which in turn results in larger gradient variance for the prior. This effect is absent in the illustrative example in \cref{fig:toy_2layer} because we evaluate the gradient variance on the same fixed model for all estimators. In \cref{app:sec:offline_eval} we compare the estimators \emph{offline} for different combinations of estimators during training. The results are in line with our online results in this section: for the gradients of the variational posterior the \dregs estimator \emph{always} has lower variance than the naive (IWAE) estimator; for the gradients of the prior the \gdregs estimator typically has lower variance, though in some cases only in the beginning of training.

\section{Related work}

\citet{Roeder2017_stl} observed that the reparameterization gradient estimator for the ELBO contains a score function term and proposed the \stl estimator %
that simply drops this term to reduce the estimator variance. They %
considered hierarchical ELBO models but do not discuss how to treat indirect score functions. While the \stl estimator is unbiased for the ELBO objective, \citet{Tucker2019_dregs} showed that it is biased for more general objectives such as the IWAE. They proposed the \dregs estimator that yields unbiased and low variance   gradients for IWAE and resolves the diminishing signal-to-noise issue of the naive IWAE gradients first discussed by \citet{Rainforth2018_bound}. We extend \dregs to hierarchical models, discuss how to treat the indirect score functions, and generalize it to general score functions by introducing \gdregs.

Several classic techniques from the variance reduction literature have been applied to variational inference and reparameterization. For example, \citet{miller2017reducing} and \citet{geffner2020approximation} proposed control variates for reparameterization gradients; \citet{Ruiz2016} used importance sampling with a proposal optimized to reduce variance. 
Such approaches are orthogonal to methods such as (G)\dregs and \stl, and can be combined with them for greater variance reduction \citep{Agrawal2020_advances_blackbox_vi}.

\section{Conclusion}

In this paper we generalized the recently proposed doubly-reparameterized gradients (\dregs, \citet{Tucker2019_dregs}) estimator for variational objectives in two ways. First, we showed that for hierarchical models such as VAEs seemingly pathwise gradients can actually contain score functions, and how to consistently and effectively extend \dregs in this case. Second, we introduced \gdregs, a doubly-reparameterized gradient estimator that applies to general score functions, while \dregs is limited to score functions of the variational distribution. Finally, we demonstrated that both generalizations can improve performance on conditional and unconditional image modelling tasks.

While we present and discuss the \gdregs estimator in the context of deep probabilistic models, it applies generally to score function gradients of the form $\expect{q_{\vphi}(\vz)}{\grad{\vtheta}{\log p_{\vtheta}(\vz)}}$. Applying it to other problem settings of this type such as normalizing flows is an exciting area of future research.

\section*{Acknowledgements}
We thank Chris Maddison as well as the anonymous reviewers for feedback on the manuscript.

\bibliography{bibliography}
\bibliographystyle{icml2021}

\clearpage

\onecolumn
\appendix
\linewidth\hsize \toptitlebar {\centering
  {\Large\bfseries Appendix to \\\ourtitle \par}}
 \bottomtitlebar 
 
\makeatletter
\renewcommand*\l@section{\@dottedtocline{1}{1.5em}{2.3em}}
\makeatother

\setcounter{page}{1}
\counterwithin{figure}{section}
\counterwithin{equation}{section}

\renewcommand{\contentsname}{Contents of the Appendix}
\addtocontents{toc}{\protect\setcounter{tocdepth}{1}}
\tableofcontents

\section{Derivation of the \gdregs identity}
\label{app:sec:GDReGs}
Here, we detail the derivation of our main result, the \gdregs identity (\cref{eq:GDReGs}), which we restate here.
\begin{equation}
    \hspace*{-.25cm}\expectsym{\vz\sim\distqz}\!\left[\gz\scorepz\right] = \expectsym{\vz\sim\distqz}\!\left[\!\left(\gz\gradtd{\vz}{\log \tfrac{\distqz}{\distpz}} + \gradtd{\vz}{\gz}\right)\left.\!\!\grad{\vtheta}{\mathcal{T}_p(\vepst; \vtheta)}\right\vert_{\vepst = \mathcal{T}_p^{-1}(\vz, \vtheta)}\right] \tag{\ref{eq:GDReGs}}
\end{equation}
We can derive this identity in two ways: (1) through a reweighting correction as presented in the main paper, see \cref{app:sec:gdregs_reweighting}; (2) through a flow-like transformation of $\vz\sim\qzx$, see \cref{app:sec:gdregs_flow}.

\subsection{Derivation via reweighting}
\label{app:sec:gdregs_reweighting}
In the main paper (\cref{sec:gdregs}), we explained that we need to make the sampling path of $\vz$ depend on the parameters $\vtheta$ to replace the score function $\grad{\vtheta}{\log\distpz}$ with a pathwise derivative. At the same time, we evaluate the objective on samples $\vz$ from the variational posterior $\qzx$ during training. The derivation consists of the following three steps:
\begin{itemize}
    \item[\circled{inner sep=1pt}{\tiny \textsf{1}}] \emph{Temporarily} change the path such that it depends on $\vtheta$. We change the path by first using importance sampling reweighting to temporarily re-write the expectation, $\expect{\distqz}{\ast} = \expect{\distpz}{\frac{\distqz}{\distpz} \ast}$ (step \circled{inner sep=1pt}{\tiny \textsf{1a}}), and then by employing reparameterization on $\distpz$: $\vz = \reparampt$ with $\vepst\sim\distpepst = \mathcal{N}(0, \mathbb{I}) $ (step \circled{inner sep=1pt}{\tiny \textsf{1b}}).
    \item[\circled{inner sep=1pt}{\tiny \textsf{2}}] Perform the gradient computation and collect all the terms.
    \item[\circled{inner sep=1pt}{\tiny \textsf{3}}] Change the path back by un-doing the reparameterization and the reweighting so we can use samples $\vz\sim\qzx$ to estimate the expectation. 
\end{itemize}
\begin{align}
    & \gradtd{\vtheta}{\expect{\distqz}{\gz}}  \aboveeq{\circled{inner sep=1pt}{\small \tiny \textsf{1a}}} \gradtd{\vtheta}{\expect{\distpz}{\tfrac{\distqz}{\distpz}\gz}} 
    \aboveeq{\circled{inner sep=1pt}{\small \tiny \textsf{1b}}} \gradtd{\vtheta}{\expect{\distpepst}{\tfrac{\distq(\reparampt)}{\distp(\reparampt)} g_{\vphi, \vtheta}\!\left(\reparampt\right)}} \label{app:eq:gdregs_reweight_lhs}\\
    & \quad \aboveeq{\circled{inner sep=1pt}{\small \tiny \textsf{2}}} \expectsym{\distpepst}\Big[   {\gradtd{\vz}{\left(\tfrac{\distqz}{\distpz}\gz\right)}\grad{\vtheta}\reparampt} + \tfrac{\distqz}{\distpz}\big(\grad{\vtheta}{\gz}-\gz\scorepz\big) \Big]_{\vz=\reparampt}\\
    & \quad \aboveeq{\circled{inner sep=1pt}{\small \tiny \textsf{3}}} \expectsym{\distqz}\Big[ \left(\gz\gradtd{\vz}{\log\tfrac{\distqz}{\distpz}}+\gradtd{\vz}{\gz}\right)\left.\grad{\vtheta}\reparampt\right\vert_{\vepst = \reparampinv} + \grad{\vtheta}{\gz} - \gz\scorepz \Big] \label{app:eq:gdregs_reweight}
\end{align}
In the derivation we have used the identity $x\grad{\ast}{\log x} = \grad{\ast}{x}$ repeatedly.
By noting that $\gradtd{\vtheta}{\expect{\distqz}{\gz}} = \expect{\distqz}{\grad{\vtheta}{\gz}}$, we can cancel these terms on the left hand side of \cref{app:eq:gdregs_reweight_lhs} and right hand side of \cref{app:eq:gdregs_reweight}. By moving $-\expect{\distqz}{\gz\scorepz}$ to the other side we obtain the desired result.

\subsection{Derivation via flow-like transformation}
\label{app:sec:gdregs_flow}

Here we provide a second derivation of the \gdregs identity (\cref{eq:GDReGs}) that does not explicitly use a re-weighting of the expectation as in \cref{app:sec:gdregs_reweighting}, but uses flow-like transformations instead \citep{Rezende2015_flows}. We already noted in the main text that we can interpret the change of paths as a flow $\vz\rightarrow\vepst\rightarrow\vz$, where $\vepst$ follows a more complicated distribution than the original $\veps$ that might have been used to reparameterize samples $\vz$ from $\qzx$. Here, we make this connection to normalizing flows more explicit.

We make use of the usual change of probability density formula for flows (see, e.g., \citet{Rezende2015_flows}):
\begin{align}
    y = f(x); x \sim p_x(x) \quad \Rightarrow \quad p_y(y) = p_x(x) \left\vert\grad{x}{f(x)}\right\vert^{-1}.
\end{align}
Our main insight is that we can reparameterize $\vz\sim\qzx$ in many different ways. Usually we sample an independent noise variable from a simple base distribution, for example, $\veps\sim\distqeps=\mathcal{N}(0,\mathbb{I})$, and use it to reparameterize $\vz$ as $\vz = \reparamq$. However, we can equally use a different base distribution $\distqtepst$ and reparameterize $\vz$ as $\vz = \reparampt, \vepst \sim \distqtepst$.   Note that we reparameterize using $\distpz$ in this case. In order for $\vz$ still to be distributed according to $\qzx$, we have to choose $\distqtepst$ to be itself given by the normalizing flow $\vepst = \reparampinv, \vz\sim\distqz$, such that $\distqtepst=\widetilde{q}(\vepst;\vphi, \vtheta)$. It may appear counter-intuitive to transform $\vz$ as $\vz\rightarrow\vepst\rightarrow\vz$ because we are doing and then un-doing a transformation; however, it allows us to  make the computation path depend on $\vtheta$ when computing the gradients. We then separate the forward flow $\vz=\reparampt$ and the backward flow $\vepst=\reparampinv$ by moving the forward flow into the path through reparameterization and the backward flow into the integration measure, see \cref{app:eq:gdregs_flow_part1} below. To move the backward flow into the integration measure, we express $\distqtepst$ as a change of density:
\begin{align}
    \vepst = \reparampinv; \vz \sim \qzx
    \quad \Rightarrow\quad 
    \distqtepst = \distqz \left\vert\grad{\vz}{\reparampinv}\right\vert^{-1}\label{app:eq:change_density_gdregs}
\end{align}
Note that $\distqtepst$ is different from $\distqeps=\mathcal{N}(0,\mathbb{I})$ typically used to reparameterize samples of the approximate posterior $\vz = \reparamq; \veps\sim\distqeps$! 
Using the change of density in \cref{app:eq:change_density_gdregs} together with reparameterization as described above, we can re-write the following expectation over $\vz$ as an integral over $\vepst$:
\begin{align}
    \int\distqz\gz\calcd{\vz} & = \int\distqtepst \gfn{\reparampt}\calcd{\vepst} && \text{reparameterization}\\
    & = \int \left[\distqz \left\vert\grad{\vz}{\reparampinv}\right\vert^{-1}\right]_{\vz=\reparampt} \gfn{\reparampt}\calcd{\vepst} && \text{change of density} \label{app:eq:gdregs_flow_part1}
\end{align}
In the these integrals, $\vepst$ is an \emph{independent} variable; its dependence on $\vtheta$ has been moved into the path ($\gfn{\reparampt}$) as well as the change of density (\cref{app:eq:change_density_gdregs}). This is one way to understand where the \verb|stop_gradient| in \cref{fig:resampled_as_if_from_schematic_single,fig:resampled_as_if_from_schematic} comes from.

To derive the \gdregs identity, we take the total derivative of \cref{app:eq:gdregs_flow_part1} \wrt $\vtheta$ and apply the chain- and product rule
\begin{align}
    & \gradtd{\vtheta}{\int\distqz\gz\calcd{\vz}} \aboveeq{\ref{app:eq:gdregs_flow_part1}} \gradtd{\vtheta}{\int \left[\distqz \left\vert\grad{\vz}{\reparampinv}\right\vert^{-1}\right]_{\vz=\reparampt} \gfn{\reparampt}\calcd{\vepst}} \\
    & \quad =  \int \gradtd{\vtheta}{\left[\left.\distqz \left\vert\grad{\vz}{\reparampinv}\right\vert^{-1}\gz
    \right\vert_{\vz=\reparampt}\right]} \;\calcd{\vepst}\\
    \begin{split}
         & \quad = \int \gradtd{\vz}{\left[\distqz \left\vert\grad{\vz}{\reparampinv}\right\vert^{-1}\gz
        \right]_{\vz=\reparampt}}\grad{\vtheta}{\reparampt} + {} \\
        & \qquad\qquad {} + \grad{\vtheta}{\left[\distqz \left\vert\grad{\vz}{\reparampinv}\right\vert^{-1}\gz
        \right]_{\vz=\reparampt}} \;\calcd{\vepst} 
    \end{split}\\
    \begin{split}
        & \quad = \int  \left.\left[\distqz \left\vert\grad{\vz}{\reparampinv}\right\vert^{-1}\right]\left(\gz\gradtd{\vz}{\log\left(\distqz \left\vert\grad{\vz}{\reparampinv}\right\vert^{-1}\right)} + \gradtd{\vz}{\gz}\right)\right\vert_{\vz=\reparampt}\grad{\vtheta}{\reparampt} + {}\\
        & \qquad\qquad {} + \left.\left[\distqz \left\vert\grad{\vz}{\reparampinv}\right\vert^{-1}\right]\left(\gz\grad{\vtheta}{\log\left\vert\grad{\vz}{\reparampinv}\right\vert^{-1}} + \grad{\vtheta}{\gz}\right)\right\vert_{\vz=\reparampt} \;\calcd{\vepst} 
    \end{split}\label{app:eq:gdregs_flow_part2}
\end{align}
where we have separated out the derivatives and used $x\grad{\ast}{\log x} = \grad{\ast}{x}$. We can now further separate terms and undo the change of density to replace $\distqz \left\vert\grad{\vz}{\reparampinv}\right\vert^{-1}=\distqtepst$ (\cref{app:eq:change_density_gdregs}) after taking the derivatives. We obtain
\begin{align}
    & \gradtd{\vtheta}{\int\distqz\gz\calcd{\vz}} = \\
    \begin{split}& \qquad = \int \distqtepst \left\{\left.\gradtd{\vz}{\left(\log\distqz + \log\left\vert\grad{\vz}{\reparampinv}\right\vert^{-1}\right)}\gz\right\vert_{\vz=\reparampt}\grad{\vtheta}{\reparampt}\right. + {}\\
    &\qquad\qquad\qquad + \left.\gradtd{\vz}{\gz}\right\vert_{\vz=\reparampt} \grad{\vtheta}{\reparampt} + \left.\grad{\vtheta}{\gz}\right\vert_{\vz=\reparampt} + {} \\
    & \qquad\qquad\qquad - \bigg. \left.\gz\right\vert_{\vz=\reparampt} \left.\grad{\vtheta}{\log\left\vert\grad{\vz}{\reparampinv}\right\vert}\right\vert_{\vz=\reparampt}
    \bigg\}\calcd{\vepst}.
    \end{split}\label{app:eq:gdregs_flow_part3}
\end{align}
To evaluate the gradients of the log Jacobians, $\grad{\ast}{\log\left\vert\grad{\vz}{\reparampinv}\right\vert}$, we can combine the log Jacobians with a simple base distribution $\distqepst$ to obtain $\distpz$ because
\begin{align}
    \vz = \reparampt; \vepst \sim \distqepst = \mathcal{N}(0,\mathbb{I})
    \quad \Rightarrow\quad 
    \distpz = \distqepst \left\vert\grad{\vz}{\reparampinv}\right\vert\label{app:eq:gdregs_flow_reparamp}
\end{align}
through reparameterization of $\distpz$ and by noting that
\begin{align}
    \grad{\ast}{\log(f(x) \cdot c)} = \grad{\ast}{\log(f(x))} && \text{ if $c$ is constant \wrt $\ast$}\label{app:eq:grad_log_constant}
\end{align}
and that the simple base distribution $\distqepst$ is constant \wrt all gradients:
\begin{align}
    \grad{\ast}{\log\left\vert\grad{\vz}{\reparampinv}\right\vert} \aboveeq{(\ref{app:eq:grad_log_constant})} 
    \grad{\ast}{\log\left(\left\vert\grad{\vz}{\reparampinv} \right\vert\distqepst\right)} \aboveeq{(\ref{app:eq:gdregs_flow_reparamp})} \grad{\ast}{\log\distpz}.
\end{align}
This allows us to simplify \cref{app:eq:gdregs_flow_part3} as
\begin{align}
    \begin{split} \gradtd{\vtheta}{\int\distqz\gz\calcd{\vz}} & = \int \distqz \left\{
    \left.\left(\gz\gradtd{\vz}{\log\tfrac{\distqz}{\distpz}} + \gradtd{\vz}{\gz}\right)\grad{\vtheta}{\reparampt}\right\vert_{\vepst = \reparampinv} +{}\right.\\
    &\qquad\qquad\qquad\qquad {}+ \bigg.\grad{\vtheta}\gz - \gz\grad{\vtheta}{\log\distpz}
    \bigg\}\calcd{\vz},
    \end{split}\label{app:eq:gdregs_flow_part4}
\end{align}
which is identical to \cref{app:eq:gdregs_reweight} and yields the \gdregs identity as explained above.

\section{Derivation of the \gdregs estimator for the IWAE objective}
\label{app:sec:gdregs_iwae}
In this section we apply the \gdregs identity derived above to derive the \gdregs estimator for the IWAE objective, \cref{eq:IWAE_GDReGs} in the main paper.

\subsection{Preliminaries on the IWAE objective}
The importance weighted autoencoder (IWAE) objective is given by
\begin{align}
     \liwae & = \expect{\vz_{1:K}\sim\qzkx}{\log\left(\frac{1}{K}\sum_{k=1}^K w_k\right)} && w_k = \frac{\distpz p(\vx \given \vz_k)}{\qzkx} \label{app:eq:iwae_objective}
\end{align}
where $\wk$ are the importance weights \citep{Burda2015_iwae}.

Due to the structure of the IWAE objective, any gradient \wrt any of its parameters can be written as 
\begin{align}
    \gradtd{\ast}{\liwae} & = \expect{\veps_{1:K}\sim\distqeps}{\sum_{k=1}^K \normwk \gradtd{\ast}{\log\wk}}; && \normwk = \frac{w_k}{\sum_j w_j}
    \label{app:eq:grad_iwae_general}
\end{align}
using the chain rule and $\grad{\ast}{\wk} = \wk\grad{\ast}{\log\wk}$. \normwk are the normalized importance weights, and we have reparameterized $\vz_k$ as $\reparamqk$. Typically, the derivatives we are interested in are \wrt the parameters \vphi and \vtheta. 

We also note the following identity that we use in the derivation of the doubly reparameterized estimators,
\begin{align}
    \gradtd{\vz}{\normwk} = \left(\normwk - \normwk^2\right)\gradtd{\vz}{\log\wk} \label{app:eq:normwk_derivative}
\end{align}
which follows from applying the chain-rule and using $\grad{\ast}{\wk} = \wk\grad{\ast}{\log\wk}$.

\citet{Tucker2019_dregs} derive the \dregs identity (\cref{eq:dregs_eps}) and use it to derive the following doubly-reparameterized gradient estimator (\dregs) \wrt the approximate posterior parameters $\vphi$ as:
\begin{align}
    \dregsgrad{\vphi}{\mathcal{L}_\text{IWAE}} & = \sum_{k=1}^K \normwk^2 \gradtd{\vz_k}{\log w_k} \gradtd{\vphi}{\reparamqk}. && \veps_{1:K}\sim\distqeps
\end{align}

\subsection{Derivation of the \gdregs estimator}

Similarly, we can derive a generalized doubly-reparameterized gradient (\gdregs) estimator \wrt the prior parameters $\vtheta$. We use the \gdregs identity (\cref{eq:GDReGs}) derived above with $\gz = \normwk$
and note that the reweighting term $\log\frac{q_{\vphi}(z)}{p_{\vtheta}(z)}$ looks like a log importance weight except for the missing likelihood:
\begin{align}
    & \gradtd{\vtheta}\liwae = \expect{\vz_{1:K}\sim\qzkx}{\sum_{k=1}^K \normwk \gradtd{\vtheta}{\log\wk}} = \expect{\vz_{1:K}\sim\qzkx}{\sum_{k=1}^K \normwk \gradtd{\vtheta}{\lpz}} \\
    & \qquad \aboveeq{(\ref{eq:GDReGs})} \expect{\vz_{1:K}\sim\qzkx}{\sum_{k=1}^K \left(\normwk\gradtd{\vz_k}{\log \frac{\qzkx}{\pzk}} + 
    \gradtd{\vz_k}{\normwk}\right)\left.\grad{\vtheta}{\reparampkt}\right\vert_{\vepst_k=\reparampkinv}} \\
    & \qquad \aboveeq{(\protect\ref{app:eq:normwk_derivative})} \expect{\vz_{1:K}\sim\qzkx}{\sum_{k=1}^K \left(\normwk\gradtd{\vzk}{\log \frac{\qzkx}{\pzk}} + 
    \left(\normwk - \normwk^2\right)\gradtd{\vzk}{\log\wk}\right)\left.\grad{\vtheta}{\reparampkt}\right\vert_{\vepst_k=\reparampkinv}} \\
    & \qquad = \expect{\vz_{1:K}\sim\qzkx}{\sum_{k=1}^K \left(\normwk  \gradtd{\vzk}{\log\pxzk} - \normwk^2 \gradtd{\vz}{\log w_k}\right) \left.\grad{\vtheta}{\reparampkt}\right\vert_{\vepst_k=\reparampkinv}}. 
\end{align}
Thus, the \gdregs estimator is given by:
\begin{align}
    \gdregsgrad{\vtheta}{\liwae} & = \sum_{k=1}^K \left(\normwk  \gradtd{\vzk}{\log\pxzk} - \normwk^2 \gradtd{\vz}{\log w_k}\right) \left.\grad{\vtheta}{\reparampkt}\right\vert_{\vepst_k=\reparampkinv} && \vz_{1:K}\sim\qzkx. \tag{\ref{eq:IWAE_GDReGs}}
\end{align}
Note that the $\vz_k$ are sampled from $\qzkx$ but re-rexpressed as if they came from $\distpz$. 

We can rewrite the importance weights as 
\begin{equation}
    \wk = \frac{\pzk \pxzk}{\qzkx} = \frac{p_{\vtheta}(\vz_k \given \vx) p_{\vtheta}(\vx)}{q_{\vphi}(\vz_k \given \vx)}.
\end{equation}
Thus, if the variational posterior $\qzkx$ is equal to the true posterior $p_{\vtheta}(\vz_k \given \vx)$, all weights $\wk$ become equal to $p_{\vtheta}(\vx)$ and thus constant \wrt $\vzk$. In that case the second term in the \gdregs estimator \cref{eq:IWAE_GDReGs} vanishes and the overall expression simplifies to
\begin{align}
    \gdregsgrad{\vtheta}{\liwae}
        = \sum_{k=1}^K \normwk \gradtd{\vzk}{\log p(\vx \given \vz_k)}\left.\grad{\vtheta}{\reparampkt}\right\vert_{\vepst_k=\reparampkinv}. && \vz_{1:K}\sim\qzkx \label{app:eq:gdregs_exact_posterior}
\end{align}
In contrast, the usual IWAE gradient involves the score function for $\pzk$:
\begin{align}
    \naivegrad{\vtheta}{\liwae}
        =  \sum_{i=1}^K \normwk  \grad{\vtheta}{\log \pzk}, && \vz_{1:K}\sim\qzkx.
\end{align}
\clearpage 

\section{Derivation of the \dregs and \gdregs estimator for IWAE objectives of hierarchical VAEs}
\label{app:sec:hierarchical_dregs_gdregs}
In this section we derivations of and further details on the extension of \dregs and \gdregs to hierarchical VAEs with the IWAE objective. 

\subsection{Preliminaries and notation for the hierarchical IWAE objective}

For a hierarchically structured model with $L$ stochastic layers the IWAE objective is still given by \cref{app:eq:iwae_objective} but with importance weights \wk given by 
\begin{align}
    \wk & = \frac{p_{\vlambda}(\vx\given\vz_{k1}, \dots, \vz_{kL}) \; p_{\vtheta}(\vz_{k1}, \dots, \vz_{kL})}{q_{\vphi}(\vz_{k1}, \dots, \vz_{kL}\given \vx)}.
\end{align}
Here, $\vz_{kl}$ denotes the $k$th importance sample ($k\in\{1,\dots, K\}$) for the $l$th layer ($l\in\{1,\dots,L\}$). Both the variational posterior and the prior distribution factorize according to their respective hierarchical structure. While the prior factorizes top-down in most cases, the variational posterior can have many different structures. In order for the distributions to be valid in the context of a VAE, we require the individual dependency graphs for the prior (generative path) and the variational posterior (inference path) to be directed acyclic graphs. Cycles would mean that a latent variable conditionally dependent on itself. To keep the dependency structure general, we write the factorization of the variational posterior and prior as follows:
\begin{align}
    q_{\vphi}(\vz_{k1}, \dots, \vz_{kL}\given \vx) & = \prod_{l=1}^L q_{\vphi_l}(\zkl\given\pa{\valpha}{l}, \vx) = 
    \prod_{l=1}^L q_{\valpha_l(\pa{\valpha}{l}; \vphi_l)}(\zkl) \\
    p_{\vtheta}(\vz_{k1}, \dots, \vz_{kL}) & = \prod_{l=1}^L p_{\vtheta_l}(\zkl\given\pa{\vbeta}{l}) = \prod_{l=1}^L p_{\vbeta_l(\pa{\vbeta}{l}; \vtheta_l)}(\zkl)
\end{align}
Here, $\valpha_l(\cdot; \vphi_l)$ and $\vbeta_l(\cdot; \vtheta_l)$ are the distribution parameters of the variational posterior and prior distribution in the $l$th layer, respectively, and we have made the dependencies of the conditional distributions explicit; $\pa{\valpha}{l}$ denotes the ``parents'' of the latent variable $\zkl$ according to the dependency graph of the inference path (the factorization of the posterior); similarly, $\pa{\vbeta}{l}$ denotes the latent variables that $\zkl$ directly depends on according to the factorization of the prior $p_{\vtheta}$. Typically, the prior is assumed to factorize top-down, such that $\pa{\vbeta}{l} = \vz_{k(l+1)}$ for all but the top-most layer.

The samples $\vz_{kl}$ are drawn from the variational posterior and can be expressed through reparameterization as $\vz_{kl} = \mathcal{T}_{q_l}(\veps_{kl}; \valpha_l(\pa{\valpha}{l}, \vphi_{l}))$, where $\epskl$ is an independent noise variable per importance sample and layer. %

We note that it is these dependencies of the distribution parameters $\valpha_l$ and $\vbeta_l$ on $\pa{\valpha}{l}$ and $\pa{\vbeta}{l}$, respectively, that give rise to the indirect score functions as discussed in \cref{sec:dregs_hierarchical}.

\subsection{Derivation of the hierarchical \dregs estimator for IWAE}

With notation fully set up we consider the reparameterized gradients of the IWAE objective \wrt the variational parameters in a particular stochastic layer $\vphi_l$:
\begin{align}
    \gradtd{\vphi_l}{\liwae} & = \expect{\veps_{1:K}\sim\distqeps}{\sum_{k=1}^K \normwk\gradtd{\vphi_l}{\log\wk}} \\
    & = \expect{\veps_{1:K}\sim\distqeps}{\sum_{k=1}^K \normwk\left(\gradtd{\zkl}{\log\wk}\grad{\vphi_l}{\reparam{q_l}{\epskl; \valpha_l(\pa{\valpha}{l}, \vphi_l)}} + \grad{\vphi_l}{\log\wk}\right)} \label{app:eq:dregs_hierarchical_part1}
\end{align}
where we have used the chain-rule to arrive at \cref{app:eq:dregs_hierarchical_part1}; the first term contains both the (true) pathwise gradients as well as the indirect score functions; the second term only contains a direct score function as we only take the partial derivative \wrt $\vphi_l$. 

We can rewrite this direct score function gradient because only one term in the (log-)importance weight directly depends on $\vphi_l$,
\begin{align}
    \grad{\vphi_l}{\log\wk} & = -\grad{\vphi_l}{\log q_{\valpha_l(\pa{\valpha}{l}; \vphi_l)}(\zkl)}.
\end{align}
Applying the \dregs identity to this term and using \cref{app:eq:normwk_derivative} yields:
\begin{align}
    \expect{\veps_{1:K}\sim\distqeps}{\sum_{k=1}^K \normwk \grad{\vphi_l}{\log\wk}}
    & = -\expect{\veps_{1:K}\sim\distqeps}{\sum_{k=1}^K (\normwk-\normwk^2)\gradtd{\zkl}{\log\wk}\grad{\vphi_l}{\reparam{q_l}{\epskl; \valpha_l(\pa{\valpha}{l}, \vphi_l)}}}
\end{align}
which agrees with the first term in \cref{app:eq:dregs_hierarchical_part1} up to the prefactor. Thus, both the true pathwise gradients as well as the indirect score functions appear twice and the prefactors partly cancel to give rise to the \dregs estimator for hierarchical IWAE objectives:
\begin{tcolorbox}[enhanced,colback=white,%
    colframe=C1!75!black, attach boxed title to top right={yshift=-\tcboxedtitleheight/2, xshift=-.75cm}, title=\dregs estimator for hierarchical IWAE objectives, coltitle=C1!75!black, boxed title style={size=small,colback=white,opacityback=1, opacityframe=0}, size=title, enlarge top initially by=-\tcboxedtitleheight/2, left=-5pt]
 \vskip-1.em
\begin{align}
    \dregsgrad{\vphi_l}{\liwae} & = \sum_{k=1}^K \normwk^2\gradtd{\zkl}{\log\wk}\grad{\vphi_l}{\reparam{q_l}{\epskl; \valpha_l(\pa{\valpha}{l}, \vphi_l)}} && \veps_{1:K}\sim\distqeps \label{app:eq:iwae_hierarchical_dregs}
\end{align}
\end{tcolorbox}
where $\vz_{kl} = \mathcal{T}_{q_l}(\veps_{kl}; \valpha_l(\pa{\valpha}{l}, \vphi_{l})), \forall l\in\{1, \dots, L\}, \forall k\in\{1, \dots, K\}$ through reparameterization. 

We emphasize that the total derivative \wrt $\zkl$ contains pathwise gradients as well as indirect score functions for both the variational posterior as well as for the prior. The hierarhical \dregs estimator otherwise looks very similar to the \dregs estimator in the single layer case \citep{Tucker2019_dregs}.

In \cref{app:sec:objectives_dregs} we explain how to implement this estimator effectively and in a structure-agnostic way. That is, we do \emph{not} have to derive a new estimator for each new dependency graph of the variational posterior or the prior.

\subsection{Derivation of the hierarchical \gdregs estimator for IWAE}
Next, we derive the expression for the \gdregs estimator for hierarchical VAEs with IWAE objective.

Applying the \gdregs identity entails re-expressing the samples $\zkl$ from the variational posterior as if they were sampled from the prior. Starting form a sample $(\vz_{k1}, \dots, \vz_{kL}) \sim q_{\vphi}(\vz_1, \dots, \vz_L\given\vx)$, we use the inverse flow of $p_{\vtheta}$ to obtain new noise variables for each layer, $(\vepst_{k1}, \dots, \vepst_{kL})$. We then use the forward flow of $p_{\vtheta}$ to obtain back $(\vz_{k1}, \dots, \vz_{kL})$ but with the gradient path now depending on $\vtheta$ as discussed in \cref{app:sec:GDReGs,sec:gdregs}. 

More precisely, we find that
\begin{align}
    \zkl^{(q)} & = \reparam{q_l}{\epskl; \valpha_l(\pa{\valpha}{l}, \vphi_l)} && \text{original sampling of $(\vz_{k1}, \dots, \vz_{kL}) \sim q_{\vphi}(\vz_1, \dots, \vz_L\given\vx)$}\\
    \epstkl & = \reparaminv{p_l}{\zkl^{(q)}; \vbeta_l(\pa{\vbeta}{l}, \vtheta_l)} && \text{inverse prior flow to obtain new ``noise'' variables}\\
    \zkl & = \reparam{p_l}{\epstkl; \vbeta_l(\pa{\vbeta}{l}, \vtheta_l)} && \text{forward prior flow to re-express the $\zkl$}
\end{align}
where $\epskl\sim\distqeps$ follows a simple distribution that is different from the more complicated distribution of $\epstkl$. Note how the initial reparameterization of a sample $\zkl$ depends on the dependency structure of the variational posterior (through $\pa{\valpha}{\cdot}$), while the other transformations depend on the dependency structure of the prior ($\pa{\vbeta}{\cdot}$). 

As for \dregs, we note that only one term in the log importance weight directly depends on the variable $\vtheta_l$,
\begin{align}
    \grad{\vtheta_l}{\log\wk} & = \grad{\vtheta_l}{\log p_{\vbeta_l(\pa{\vbeta}{l}; \vtheta_l)}(\zkl)}.  \label{app:eq:grad_theta_logw}
\end{align}
With these prerequesits, we can compute the \gdregs estimator for parameters $\vtheta_l$ of the $l$th stochastic layer. 
\begin{align}
    \gradtd{\vtheta_l}{\liwae} & = \expect{\vz_{1:K}\sim\qzx}{\sum_{k=1}^K \normwk\grad{\vtheta_l}{\log\wk}} \\
    & \aboveeq{\ref{app:eq:grad_theta_logw}} \expect{\vz_{1:K}\sim\qzx}{\sum_{k=1}^K\normwk\grad{\vtheta_l}{\log p_{\vbeta_l(\pa{\vbeta}{l}; \vtheta_l)}(\zkl)}} \\
    \begin{split}
    &  \aboveeq{\ref{eq:GDReGs}} \expectsym{\vz_{1:K}\sim\qzx}\bigg[\sum_{k=1}^K \Big(\normwk\gradtd{\zkl}{\log \frac{q_{\vphi}(\vz_{k1}, \dots, \vz_{kL}\given\vx)}{p_{\vtheta}(\vz_{k1}, \dots, \vz_{kL})}} + {} \Big.\bigg.\\
    & \qquad\qquad \qquad\qquad\qquad\qquad \bigg.\Big.{}+ \left(\normwk - \normwk^2\right)\gradtd{\zkl}{\log\wk}\Big)\left.\grad{\vtheta_l}{\reparam{p_l}{\epstkl; \vtheta_l}}\right\vert_{\epstkl=\reparaminv{p_l}{\zkl; \vtheta_l}}\bigg]
    \end{split}
\end{align}
\begin{tcolorbox}[enhanced,colback=white,%
    colframe=C1!75!black, attach boxed title to top right={yshift=-\tcboxedtitleheight/2, xshift=-.75cm}, title=\gdregs estimator for hierarchical IWAE objectives, coltitle=C1!75!black, boxed title style={size=small,colback=white,opacityback=1, opacityframe=0}, size=title, enlarge top initially by=-\tcboxedtitleheight/2, left=-5pt]
 \vskip-1.em
\begin{align}
    & \gdregsgrad{\vtheta_l}{\liwae} = \label{app:eq:iwae_hierarchical_gdregs}  \\
    & \quad = \sum_{k=1}^K \left(\normwk  \gradtd{\zkl}{\log p_{\vlambda}(\vx\given\vz_{k1}, \dots, \vz_{kL})} - \normwk^2 \gradtd{\zkl}{\log w_k}\right) \left.\grad{\vtheta_l}{\reparam{p_l}{\epstkl; \vtheta_l}}\right\vert_{\epstkl=\reparaminv{p_l}{\zkl; \vtheta_l}}; && \vz_{1:K}\sim\qzkx \nonumber
\end{align}
\end{tcolorbox}
where we suppressed dependencies on $\pa{\ast}{l}$ where they are not necessary to simplify notation.

The estimator looks very similar to the \gdregs estimator for a single layer IWAE model \cref{eq:IWAE_GDReGs}. Note that just like above for hierarchical \dregs, the total gradients \wrt $\zkl$ give rise to both (true) pathwise gradients as well as indirect score functions through the hierarchical dependencies of the variational posterior and prior. 

In \cref{app:sec:surr_loss_gdregs} we show how to implement the hierarchical \gdregs estimator \cref{app:eq:iwae_hierarchical_gdregs} effectively and in a way that is agnostic to the structure of the model. That is, we do not have to derive a separate estimator for every dependency graph of the variational posterior and prior.

\subsection{Double reparameterization and indirect score functions}
\label{app:sec:double_reparameterization_indirect_score}
In principle, we could apply double-reparameterization to the indirect score functions as well. However, as we explain now, we often cannot doubly-reparameterize \emph{all} indirect score functions; moreover, even in cases where this is possible, it is still impractical, as the corresponding estimator depends on the exact model structure and would require adaptation to each dependency graph of the prior and variational posterior. 

Double reparameterization of indirect score functions works in the same way as for the direct score functions except that $g_{\vphi, \vtheta}(\vz)$ is given by $\normwk^2$ instead of $\normwk$ in this case. The derivatives of $\normwk^2$ have a similar reproducing property as we observed in \cref{app:eq:normwk_derivative}:
\begin{align}
    \gradtd{\vz}{\normwk^2} & = 2(\normwk^2-\normwk^3)\gradtd{\vz}{\log\wk}.
\end{align}
Thus, double reparameterization of the indirect score functions similarly gives rise to further indirect score functions. We note that these indirect score functions only appear for the ``children'' of the current stochastic layer, that is, stochastic variables in those layers that depend on the current layer. In this context, ``children'' refers to \emph{all} children \wrt the dependency structure of both, the variational posterior \emph{and} the prior. For a particular layer $l$ we obtain indirect score functions from double reparameterization of all of its (direct or indirect) parent nodes. Following the dependency structure, we could collect all of these terms and reparameterize them to obtain pathwise gradients only. 

However, a problem arises, because we need to account for dependencies of both the variational posterior \emph{and} the prior. Reparameterization of a score function gives rise to indirect score functions in all its ``children'' layers for both the variational posterior and the prior. For general hierarchical structures, this leads to cycles, in that some of the children of one dependency tree (the variational posterior) are the parents in the other (the prior) and/or vice versa. In this case we are never able to collect all the terms and fully reparameterize all the score functions.

Moreover, even if the joint dependency graph of the variational posterior and the prior were acyclic, this derivation would be structure-specific and would need to be repeated for each hierarchical structure. 
We therefore do not doubly reparameterize the indirect score functions.

\clearpage

\section{Worked example for a 2-layer hierarchical VAE}
\label{app:sec:worked_hierarchical_example}
In this section we show an example of using the IWAE objective with a $2$-layer VAE model consisting of a prior $p_{\vtheta_2}(\vz_2)p_{\vtheta_1}(\vz_1\given \vz_2)$ and likelihood $p_{\vlambda}(\vx\given\vz_1, \vz_2)$. The inference network is bottom-up: $q_{\vphi_1}(\vz_1\given\vx)q_{\vphi_2}(\vz_2\given\vx, \vz_1)$. Therefore $\pa{\valpha}{2} = \vz_1$ and $\pa{\vbeta}{1} = \vz_2$.

We hierarchically sample $\vz_{k1}$ and $\vz_{k2}$ from the approximate posterior and abbreviate: 
\begin{align}
    \vz_{k1}(\vphi_1) &\equiv \mathcal{T}_{q_1}(\veps_{k1}; \valpha_1(\vphi_1))\\
    \vz_{k2}(\vphi_1, \vphi_2) &\equiv \mathcal{T}_{q_{2\given 1}}(\veps_1; \valpha_{2\given 1}(\vx, \vz_{k1}(\vphi_1), \vphi_2)).
\end{align}
 We also explicitly distinguish between the distribution parameters $\valpha_i$ and the network parameters $\vphi_i$ for the posterior as well as the distribution parameters $\vbeta_i$ and the network parameters $\vtheta_i$ for the prior. A single importance sample \wk with all of its functional dependencies is given by:
\begin{align}
    \wk & = \frac{\plambda\left(\vx|\vz_{k1}(\vphi_1), \vz_{k2}(\vphi_1, \vphi_2)\right) \;\; p_{\vbeta_2(\vtheta_2)}\left(\vz_{k2}(\vphi_1, \vphi_2)\right) \;\; p_{\vbeta_{1\given 2}(\vz_{k2}(\vphi_1, \vphi_2), \vtheta_1)}\left(\vz_{k1}(\vphi_1)\right)}{q_{\valpha_1(\vx,\vphi_1)}(\vz_{k1}(\vphi_1)) \;\; q_{\valpha_{2\given 1}(\vx,\vz_{k1}(\vphi_1), \vphi_2)}\left(\vz_{k2}(\vphi_1, \vphi_2)\right)}.
\end{align}
The \dregs estimator for the variational parameters of the lower latent layer, $\vphi_1$, is then given by:
\begin{align}
    \dregsgrad{\vphi_1}{\liwae} & = \sum_{k=1}^K \normwk^2\gradtd{\vz_{k1}}{\log\wk}\grad{\vphi_1}{\reparam{q_1}{\veps_{k1}; \valpha_1(\vx, \vphi_1)}}; \qquad\qquad \veps_{1:K}\sim\distqeps \\
    \begin{split}
    \gradtd{\vz_{k1}}{\log\wk} & = \grad{\vz_{k1}}{\log\wk} - \grad{\valpha_{2\given 1}}{\log q_{\valpha_{2\given 1}(\vx,\vz_1(\vphi_1), \vphi_2)}\left(\vz_{k2}(\vphi_1, \vphi_2)\right)}\grad{\vz_{k1}}{\valpha_{2\given 1}(\vx,\vz_{k1}(\vphi_1), \vphi_2)} \\
    & \qquad \qquad \qquad + \grad{\vbeta_{1 \given 2}}{\log p_{\vbeta_{1\given 2}(\vz_{k2}(\vphi_1, \vphi_2), \vtheta_1)}\left(\vz_{k1}(\vphi_1)\right)} \grad{\vz_{k1}}{\vbeta_{1\given 2}(\vz_{k2}(\vphi_1, \vphi_2), \vtheta_1)}
    \end{split}
\end{align}
where we have expanded the total derivative \wrt $\vz_{k1}$ into the (true) pathwise gradients and two indirect score functions.

Similarly, we can compute the \dregs estimator for the variational parameters of the upper level, $\vphi_2$:
\begin{align}
    \dregsgrad{\vphi_2}{\liwae} & = \sum_{k=1}^K \normwk^2\gradtd{\vz_{k2}}{\log\wk}\grad{\vphi_2}{\reparam{q_{2\given 1}}{\veps_{k2}; \valpha_2(\vx, \vz_{k1}(\vphi_1), \vphi_2)}}; \qquad\qquad \veps_{1:K}\sim\distqeps \\
    \gradtd{\vz_{k2}}{\log\wk} & = \grad{\vz_{k2}}{\log\wk} + \grad{\vbeta_{1 \given 2}}{\log p_{\vbeta_{1\given 2}(\vz_2(\vphi_1, \vphi_2), \vtheta_1)}\left[\vz_1(\vphi_1)\right]} \grad{\vz_{k2}}{\vbeta_{1\given 2}(\vz_2(\vphi_1, \vphi_2), \vtheta_1)}
\end{align}
For this model structure of the prior and variational posterior, there is only one indirect score function for this gradient. Note that the indirect score functions are computed automatically in our surrogate losses that we introduce in the following section, such that we do not need to compute them manually; the hierarchical \dregs estimator can be implemented without having to trace the dependency structure of the model manually.

To compute the \gdregs estimator, we first have to re-express the samples $\vz_1$ and $\vz_2$ as if they were sampled from $p_{\vtheta_1, \vtheta_2}(\vz_1, \vz_2)$. We write this reparameterization as
\begin{align}
    \vz_{k2}(\vtheta_2) & \equiv \reparam{p_2}{\vepst_{k2}; \vbeta_2(\vtheta_2)} \\
    \vz_{k1}(\vtheta_1, \vtheta_2) & \equiv \reparam{p_{2\given 1}}{\vepst_{k1}; \vbeta_{1\given 2}(\vz_{k2}(\vtheta_2), \vtheta_1)}
\end{align}
where $\vepst_1$ and $\vepst_2$ are noise variables drawn from $\distqtepst$, which is given by the inverse prior flow of the samples drawn from $q_{\vphi_1, \vphi_2}(\vx_1, \vx_2\given\vx)$. The full functional dependency of a single importance sample is given by:
\begin{align}
    \wk & = \frac{\plambda\left(\vx|\vz_{k1}(\vtheta_1, \vtheta_2), \vz_{k2}(\vtheta_2)\right) \;\; p_{\vbeta_2(\vtheta_2)}\left(\vz_{k2}(\vtheta_2)\right) \;\; p_{\vbeta_{1\given 2}(\vz_{k2}(\vtheta_2), \vtheta_1)}\left(\vz_{k1}(\vtheta_1, \vtheta_2)\right)}{q_{\valpha_1(\vx,\vphi_1)}(\vz_{k1}(\vtheta_1, \vtheta_2)) \;\; q_{\valpha_{2\given 1}(\vx,\vz_{k1}(\vtheta_1, \vtheta_2), \vphi_2)}\left(\vz_{k2}(\vtheta_2)\right)}.
\end{align}

The \gdregs estimator \wrt the prior parameters of the lower stochastic layer, $\vtheta_1$, is given by:
\begin{align}
    \gdregsgrad{\vtheta_1}{\liwae} & = \sum_{k=1}^K \left(\normwk \gradtd{\vz_{k1}}{\log p_{\vlambda}(\vx\given\vz_{k1}, \vz_{k2})} - \normwk^2 \gradtd{\vz_{k1}}{\log w_k}\right) \left.\grad{\vtheta_1}{\reparam{p_1}{\vepst_{k1}; \vtheta_1}}\right\vert_{\vepst_{k1}=\reparaminv{p_1}{\vz_{k1}; \vtheta_1}}; \qquad \vz_{1:K}\sim\qzkx \nonumber\\
    \gradtd{\vz_{k1}}{\log w_k} & = \grad{\vz_{k1}}{\log w_k} - \grad{\valpha_{2\given 1}}{\log q_{\valpha_{2\given 1}(\vx,\vz_{k1}(\vtheta_1, \vtheta_2), \vphi_2)}\left(\vz_{k2}(\vtheta_2)\right)}\grad{\vz_{k1}}{\valpha_{2\given 1}(\vx,\vz_{k1}(\vtheta_1, \vtheta_2), \vphi_2)} 
\end{align}

The \gdregs estimator \wrt the prior parameters of the upper stochastic layer, $\vtheta_2$, is given by:
\begin{align}
    \gdregsgrad{\vtheta_2}{\liwae} & = \sum_{k=1}^K \left(\normwk \gradtd{\vz_{k2}}{\log p_{\vlambda}(\vx\given\vz_{k1}, \vz_{k2})} - \normwk^2 \gradtd{\vz_{k2}}{\log w_k}\right) \left.\grad{\vtheta_2}{\reparam{p_2}{\vepst_{k2}; \vtheta_2}}\right\vert_{\vepst_{k2}=\reparaminv{p_2}{\vz_{k2}; \vtheta_2}}; \qquad \vz_{1:K}\sim\qzkx \nonumber\\
    \begin{split}
    \gradtd{\vz_{k2}}{\log w_k} & = \grad{\vz_{k2}}{\log w_k} - \grad{\valpha_{2\given 1}}{\log q_{\valpha_{2\given 1}(\vx,\vz_{k1}(\vtheta_1, \vtheta_2), \vphi_2)}\left(\vz_{k2}(\vtheta_2)\right)}\grad{\vz_{k2}}{\valpha_{2\given 1}(\vx,\vz_1(\vtheta_1, \vtheta_2), \vphi_2)} \\
    & \qquad \qquad \qquad + \grad{\vbeta_{1 \given 2}}{\log p_{\vbeta_{1\given 2}(\vz_2(\vtheta_2), \vtheta_1)}\left(\vz_{k1}(\vtheta_1, \vtheta_2)\right)} \grad{\vz_{k2}}{\vbeta_{1\given 2}(\vz_{k2}(\vtheta_2), \vtheta_1)}
    \end{split}
\end{align}
Note how the \gdregs estimator for $\vtheta_2$ has two indirect score functions for the upper layer where the \dregs estimator for $\vphi_2$ only has one. This is due to the opposite factorization (opposite hierarchical dependency structure) of the variational posterior and the prior. This cyclic dependence is also the reason why we cannot replace all indirect score functions with \dregs and \gdregs gradients. Double reparameterization of one of the indirect score functions, leads to another indirect score function, whose double-reparameterization in turn leads back to the first indirect score function but with a different pre-factor.

Again, note that the indirect score functions are computed automatically in our surrogate losses, \cref{app:sec:objectives}, and we do not need to manually trace the dependency structure or derive them.

\section{Surrogate losses to implement the \dregs and \gdregs estimators for IWAE objectives}
\label{app:sec:objectives}
As we discussed in \cref{sec:gdregs_iwae} and similar to \citet{Tucker2019_dregs}, we use surrogate loss functions to compute the gradients \wrt the likelihood, proposal, and prior parameters. That is, we use different losses, such that backpropagation results in the respective gradient estimator. While \citet{Tucker2019_dregs} use a single surrogate loss to compute the gradient estimators for all parts of the objective, we choose to use separate surrogate losses for each of the three parameter groups (likelihood, variational posterior, prior). In principle, we could combine them into a single loss, but in order to keep presentation simple we keep them separate. Computationally this does not make a difference as modern deep learning frameworks avoid duplicate computation.

For the likelihood parameters, we use the regular (negative) IWAE objective \cref{eq:iwae} as a loss. That is, the gradient estimator for the likelihood parameters is given by the gradient of the negative IWAE objective.

To construct the other surrogate losses we need to stop the gradients at various points in the computation graph. In the following, we use the shorthand notation \gradno{\protect\phantom{hello}} to indicate that we stop gradients into the underlined parts of an expression. Where it might be ambiguous, or to highlight where we do \emph{not} stop gradients, we use the shorthand \gradyes{\protect\phantom{hello}} to indicate that gradients flow. For example, $f(\gradyesm{\vphi}, \gradnom{\vtheta})$ means that we backpropagate gradients into $\vphi$ but not into $\vtheta$.

\subsection[]{\dregs for variational posterior parameters $\vphi$}
\label{app:sec:objectives_dregs}
\subsubsection{Single stochastic layer}
Here we reproduce part of the surrogate loss for the variational parameters $\vphi$ by \citet{Tucker2019_dregs} for the single stochastic layer case: 
\begin{equation}
\begin{aligned}
    L_\text{\dregs}(\vphi) & = \sum_{k=1}^K \gradnom{\normwk^2} \left(\log p_{\vlambda}(\vx\given\gradyesm{\vzk}) + \log p_{\vbeta(\vtheta)}(\gradyesm{\vzk}) - \log q_{\valpha(\gradnom{\small\vphi})}(\gradyesm{\vzk})\right) \\
    \gradyesm{\vzk} & = \reparam{q}{\epsk; \gradyesm{\vphi}} \qquad\qquad \epsk\sim q(\epsk)
\end{aligned}
\end{equation}
That is, we sample $\vzk\sim\qzkx$ as usual (by reparameterizing independent noise variables $\epsk$) but stop the gradients of the parameters that parameterize the distributions when evaluating their densities, $\log q_{\valpha(\gradnom{\small\vphi})}(\gradyesm{\vzk})$. In addition we stop the gradients around the normalized importance weights $\normwk$. Differentiating $L_\text{\dregs}$ \wrt the proposal parameters $\vphi$ yields the \dregs estimator \cref{eq:IWAE_DReGs}. Note that we do not explicitly stop gradients into $\vlambda$ or $\vtheta$ because we use separate surrogate losses for those parameter groups. If we were to use a combined loss, we would potentially have to stop gradients into these parameters as well, depending on the estimator used.

To practically implement this surrogate loss, we use two copies of the variational posterior distribution. An unaltered one (no stopped gradients) to sample $\vz$ and one with gradients into the proposal parameters stopped to evaluate the log densities. The stopped gradient makes sure that we do not obtain a direct score function as we have doubly-reparameterized it.

Note that for single-stochastic-layer models we could also stop the gradients of the distribution parameters $\valpha$ instead as they only depend on $\vphi$. We emphasize that this is not possible for hierarchical models as this would eliminate the indirect score functions and thus produce potentially biased gradients.

\subsubsection{Multiple stochastic layers}
For multiple layers, the surrogate loss for the \dregs estimator \cref{app:eq:iwae_hierarchical_dregs} is given by: 
\begin{equation}
    \begin{aligned}
        L_\text{\dregs}(\vphi) & = \sum_{k=1}^K \gradnom{\normwk^2} \log\wk\\
        \log\wk & = \log p_{\vlambda}(\vx\given\gradyesm{\vz_{k1}}, \gradyesm{\dots}, \gradyesm{\vz_{kL}}) + \sum_{l=1}^L \log p_{\vbeta_l(\gradyesms{\pa{\vbeta}{l}}; \vtheta_l)}(\gradyesm{\zkl}) - \sum_{l=1}^L \log q_{\valpha_l(\gradyesms{\pa{\valpha}{l}}; \gradnoms{\vphi_l})}(\gradyesm{\zkl}) \\
        \gradyesm{\zkl} & = \reparam{q_l}{\epskl; \valpha_l(\gradyesm{\pa{\valpha}{l}}, \gradyesm{\vphi_l})} \qquad \qquad \epskl\sim q(\epskl)
    \end{aligned}
    \label{app:eq:surr_loss_dregs_hierarchical_iwae}
\end{equation}
Again, we do not explicitly stop gradients into $\vlambda$ or $\vtheta_l$ as we only take gradients \wrt $\vphi_l$. 

The indirect score functions arise due to the indirect dependence of the distribution parameters $\valpha_l(\gradyesms{\pa{\valpha}{l}}; \gradnoms{\vphi_l})$ and $\vbeta_l(\gradyesms{\pa{\vbeta}{l}}; \vtheta_l)$ on the parent latent variables $\gradyesm{\pa{\valpha}{l}}$ and $\gradyesm{\pa{\vbeta}{l}}$, respectively. Note how the former depends on the hierarchical structure of the variational posterior, whereas the latter depends on the hierarchical structure of the prior.

To implement this surrogate loss effectively, we again use two copies of the variational posterior distribution. One un-altered one (without stopped gradiends) from which we sample the individual reparameterized $\zkl$ and through which gradients can flow; we use these samples to evaluate densities at and to parameterize the distribution parameters at subsequent layers. Derivatives \wrt $\vphi_l$ will then give rise to pathwise gradients and indirect score functions. We use the second copy of the variational posterior, where we have stopped the parameters $\vphi_l$, to evaluate the density at for the log importance weights in the last summand of \cref{app:eq:surr_loss_dregs_hierarchical_iwae}.

\subsection[]{\gdregs for prior parameter $\vtheta$}
\label{app:sec:surr_loss_gdregs}
\subsubsection{Single stochastic layer}
\begin{equation}
\begin{aligned}
    L_\text{\gdregs}(\vtheta) & = \sum_{k=1}^K \gradnom{\normwk}\log p_{\vlambda}(\vx\given\gradyesm{\vzk}) - \gradnom{\normwk^2} \left(\log p_{\vlambda}(\vx\given\gradyesm{\vzk}) + \log p_{\vbeta(\gradnoms{\vtheta})}(\gradyesm{\vzk}) - \log q_{\valpha(\vphi)}(\gradyesm{\vzk})\right) \\
    \gradyesm{\vzk} & = \reparam{p}{\gradnom{\epstk}; \gradyesm{\vtheta}} \\
    \gradnom{\epstk} & = \gradnom{\reparaminv{p}{\reparam{q}{\epsk; \vphi}; \vtheta}} \qquad \qquad \epsk\sim q(\epsk)
\end{aligned}
    \label{app:eq:surr_loss_gdregs_iwae}
\end{equation}
Taking the derivative of \cref{app:eq:surr_loss_gdregs_iwae} \wrt $\vtheta$ gives rise to the \gdregs estimator for the single stochastic layer IWAE objective. As explained in \cref{sec:gdregs}, we need to re-express $\vzk$ such that its path depends on $\vtheta$. In effect, we first sample $\vzk = \reparamqk$, then compute the new noise variable $\epstk = \reparampinv$, and re-compute $\vzk = \reparampkt$. Note that we have to stop gradients into the noise variables $\epstk$ to obtain the correct gradient estimator. This explains the \verb|stop_grad| in \cref{fig:resampled_as_if_from_schematic_single}.

As above, we do not explicitly stop gradients into $\vlambda$ and $\vphi$ as we use separate losses for these parameter groups and only compute gradients of \cref{app:eq:surr_loss_gdregs_iwae} \wrt $\vtheta$.

To effectively implement this loss, we use two copies of the prior distribution. One that we implement as a normalizing flow and a second one with stopped gradients into the parameters. We then proceed as follows:
\begin{itemize}
    \item Compute the new noise variables $\epstk$ by using the inverse flow $\mathcal{T}^{-1}_p$ on the samples $\vzk$ from the variational posterior.
    \item Stop the gradients into $\epstk$.
    \item Use the forward flow $\reparampkt$ to re-compute $\vzk$ but with path dependent on $\vtheta$. These samples when derived \wrt $\vtheta$ will give rise to the pathwise gradients.
    \item Use the second copy of the prior (with stopped gradients into its parameters) to evaluate the log density at the samples $\vzk$. The stopped gradients make sure that we do not obtain the direct score function.
\end{itemize}

\subsubsection{Multiple stochastic layers}
For multiple stochastic layers the surrogate loss that gives rise to the \gdregs estimator \cref{app:eq:iwae_hierarchical_gdregs} is given by:
\begin{equation}
    \begin{aligned}
        L_\text{\gdregs}(\vtheta) & = \sum_{k=1}^K \gradnom{\normwk} \log p_{\vlambda}(\vx\given\gradyesm{\vz_{k1}}, \gradyesm{\dots}, \gradyesm{\vz_{kL}}) - \gradnom{\normwk^2} \log\wk\\
        \log\wk & = \log p_{\vlambda}(\vx\given\gradyesm{\vz_{k1}}, \gradyesm{\dots}, \gradyesm{\vz_{kL}}) + \sum_{l=1}^L \log p_{\vbeta_l(\gradyesms{\pa{\vbeta}{l}}; \gradnoms{\vtheta_l})}(\gradyesm{\zkl}) - \sum_{l=1}^L \log q_{\valpha_l(\gradyesms{\pa{\valpha}{l}}; \vphi_l)}(\gradyesm{\zkl}) \\
        \gradyesm{\zkl} & = \reparam{p_l}{\gradnom{\epstkl}; \vbeta_l(\gradyesm{\pa{\vbeta}{l}}, \gradyesm{\vtheta_l})} \\
        \gradnom{\epstkl} & = \gradnom{\reparaminv{p_l}{\zkl^{(q)}; \vbeta_l\left(\pa{\vbeta}{l}, \vtheta_l\right)}} \\
        \zkl^{(q)} & = \reparam{q_l}{\epskl; \valpha_l\left(\pa{\valpha}{l}, \vphi_l\right)} \qquad \qquad \epskl\sim q(\epskl)
    \end{aligned}
    \label{app:eq:surr_loss_gdregs_hierarchical_iwae}
\end{equation}

As for the single layer case, we need to re-express variational posterior samples $\zkl$ as if they were sampled from the prior. To obtain the correct gradients, we again have to stop gradients into the new noise variables $\epstkl$, also see \cref{fig:resampled_as_if_from_schematic}. 

As for hierarchical \dregs, the indirect score functions stem from the second and third term of $\log\wk$ and arise because the distribution parameters $\valpha_l$ and $\vbeta_l$ depend on the ``parent'' stochastic layers.

As before we use two copies of the prior distribution, one with regular gradients that is set up as a flow, and a second with stopped gradients into the parameters. This allows us to implement the \gdregs estimator regardless of the model structure.

\section{Implementation details}
\label{app:sec:code}
In our implementations we use \numpy{} \citep{Harris2020_numpy}, \texttt{JAX} \citep{jax2018}, \haiku{} \citep{haiku2020}, as well as \texttt{tensorflow probability} and \texttt{tensorflow distributions} \citep{Dillon2017_tfp_distributions}.

\paragraph{Stopping gradients.} All major frameworks allow for gradients to be stopped or interupted. For example, in \tensorflow{} \citep{abadi2016tensorflow} we can use \verb|tf.stop_gradient| and in \jax{} we can use \verb|jax.lax.stop_gradient|.
To implement stopped gradients \wrt the parameters of a distribution we use \verb|haiku.experimental.custom_getter| contexts, which allow us to manipulate the parameters before they are used to construct the respective networks; in this case we use the context to stop gradients. 

\paragraph{Re-expressing samples.} To re-express variational posterior samples as if they came from the prior, we directly implement the computation flow as it is described in e.g. \cref{fig:resampled_as_if_from_schematic_single,fig:resampled_as_if_from_schematic} and detailed in \cref{app:sec:surr_loss_gdregs}.

In the code listings \cref{app:lst:setup,app:lst:program} we provide (pseudo-)code for a simple implementation of the surrogate objectives. \cref{app:lst:setup} contains the import statements as well as the function and class definitions to create parameterized distributions that allow for 
\begin{enumerate}
    \item stopping the gradients into their parameters, and
    \item re-expressing samples using the bijector interface in \texttt{tensorflow probability}.
\end{enumerate}

In \cref{app:lst:program} we implement surrogate objectives for the naive and the \gdregs estimators of the cross-entropy. More specifically, we wish to estimate:
\begin{align}
    \gradtd{\vtheta}{\mathcal{L}^{ce}} = \gradtd{\vtheta}{\expect{\vz\sim\qzx}{\log\distpz}}.
\end{align}
The naive estimator (using a single Monte Carlo sample) is given by 
\begin{align}
    \naivegrad{\vtheta}{\mathcal{L}^{ce}} = \grad{\vtheta}{\log\distpz} \qquad \vz\sim\distqz.
\end{align}
The corresponding \gdregs estimator (again using a single MC sample) is given by \cref{eq:gdregs:x-entropy}:
\begin{align}
    \gdregsgrad{\vtheta}{\mathcal{L}^{ce}} = \grad{\vz}{\log \tfrac{\distqz}{\distpz}}\Big.\grad{\vtheta}{\mathcal{T}_p(\vepst; \vtheta)}\Big\vert_{\vepst = \mathcal{T}_p^{-1}(\vz, \vtheta)} \qquad \vz\sim\distqz.
\tag{\ref{eq:gdregs:x-entropy}}
\end{align}

\begin{lstlisting}[xleftmargin=1cm, language=Python, label=app:lst:setup, caption=Function and class definitions necessary to define the surrogate objectives for the \dregs and \gdregs estimators.]
from typing import List
import haiku as hk
import jax
import jax.lax as lax
import jax.numpy as jnp
from tensorflow_probability.substrates import jax as jtfp
jtfd = jtfp.distributions


def stop_grad_getter(next_getter, value, _):
  """A custom getter that stops gradients of parameters."""
  return lax.stop_gradient(next_getter(value))


def reparameterize_as_if_from(p: jtfd.TransformedDistribution,
                              z_q: jnp.ndarray) -> jnp.ndarray:
  """Reparameterize samples z_q as if they were sampled from p.

  Transforms: z_q -> stop_gradient(noise) -> z_q_as_if_from_p
  This transformation leaves the numerical value unchanged but alters the
    gradients.

  Args:
    p: a TransformedDistribution object
    z_q: Samples to be reparameterized.

  Returns:
    Reparameterized samples.
  """

  eps = p.bijector.inverse(z_q)
  eps = jax.lax.stop_gradient(eps)
  return p.bijector.forward(eps)
  
 
class ConditionalNormal(hk.Module):
  """A Normal distribution that is conditioned through an MLP."""

  def __init__(
      self,
      output_size: int,
      hidden_layer_sizes: List[int],
      name: str = "conditional_normal"):
    """Creates a conditional Normal distribution.

    Args:
      output_size: The dimension of the random variable.
      hidden_layer_sizes: The sizes of the hidden layers of the fully connected
        network used to condition the distribution on the inputs.
      name: The name of this distribution.
    """
    super(ConditionalNormal, self).__init__(name=name)
    self.name = name
    self.fcnet = hk.nets.MLP(
        output_sizes=hidden_layer_sizes + [2 * output_size],
        activation=jnp.tanh,
        activate_final=False,
        with_bias=True,
        name=name + "_fcnet")

  def condition(self, inputs):
    """Computes the parameters of a normal distribution based on the inputs."""
    outs = self.fcnet(inputs)
    mu, sigma = jnp.split(outs, 2, axis=-1)
    sigma = jax.nn.softplus(sigma)
    return mu, sigma

  def __call__(self, inputs, **kwargs):
    """Creates a normal distribution conditioned on the inputs."""
    # Optional `stop_gradient_params` argument stops the parameters
    # of the distribution.
    if kwargs.get("stop_gradient_params", False):
      with hk.experimental.custom_getter(stop_grad_getter):
        mu, sigma = self.condition(inputs)
    else:
      mu, sigma = self.condition(inputs)

    # Optional `as_flow` argument parameterizes the distribution as a flow
    # to have access to `Bijector.inverse` and `Bijector.forward`
    # to use with the function `reparameterize_as_if_from`
    if kwargs.get("as_flow", False):
      bijector = jtfp.bijectors.Chain(
          [jtfp.bijectors.Shift(shift=mu), jtfp.bijectors.Scale(scale=sigma)])
      base = jtfd.Normal(loc=jnp.zeros_like(mu), scale=jnp.ones_like(sigma))
      return jtfd.TransformedDistribution(
          distribution=base, bijector=bijector, name=self.name + "_flow")

    return jtfd.Normal(loc=mu, scale=sigma)
\end{lstlisting}

\begin{lstlisting}[xleftmargin=1cm, language=Python, label=app:lst:program, caption={Code to implement the surrogate objective for the naive estimator of the cross-entropy as well as for the \gdregs estimator \protect\cref{eq:gdregs:x-entropy}}. Computing derivatives w.r.t. the parameters of the prior \texttt{p} using automatic differentiation gives rise to the correct expressions for the estimators.]
# Create distributions
# Inputs `x` and context `c`
q = ConditionalNormal(x)
p = ConditionalNormal(c)
q_stop = ConditionalNormal(x, stop_gradient_params=True)
p_stop = ConditionalNormal(c, stop_gradient_params=True)
p_flow = ConditionalNormal(c, as_flow=True)

# Sample from the variational posterior
z_q = q.sample(sample_shape=[num_samples], seed=hk.next_rng_key())  # [k, bs, z]

# Reparameterize the samples from q as if they were sampled from p
z_q_as_p = reparameterize_as_if_from(p_flow, z_q)

# Cross-entropy surrogate losses
cross_entropy_naive = p.log_prob(z_q)
cross_entropy_gdregs = q_stop.log_prob(z_q_as_p) - p_stop.log_prob(z_q_as_p)
\end{lstlisting}

\section{Experimental details and additional results}
\label{app:sec:exp_details_results}

In this section we provide additional experimental details as well as additional results.

\subsection{Illustrative example}
As discussed in the main text, we use a $2$-layer linear VAE inspired by the single layer example of \citet{Rainforth2018_bound,Tucker2019_dregs}. We use a top-down generative model $\vz_2 \rightarrow\vz_1\rightarrow\vx$, with $\vz_1, \vz_2, \vx \in \mathbb{R}^D$ and $D=5$. The hierarchical prior is given by $\vz_2 \sim \mathcal{N}(0, \mathbb{I})$, $\vz_1 \given \vz_2 \sim \mathcal{N}(\vmu_{\vtheta}(\vz_2), \vsigma^2_{\vtheta}(\vz_2))$, and the likelihood is given by $\vx\given\vz_1 \sim \mathcal{N}(\vz_1, \mathbb{I})$. We choose a bottom-up variational posterior that factorizes as: $q_{\vphi_1}(\vz_1 \given \vx) = \mathcal{N}(\vmu_{\vphi_1}(\vx), \vsigma^2_{\vphi_1}(\vx))$ and $q_{\vphi_2}(\vz_2 \given \vz_1) = \mathcal{N}(\vmu_{\vphi_2}(\vz_1), \vsigma^2_{\vphi_2}(\vz_1))$. All functions $\vmu_\ast$ and $\vsigma_\ast$ are given by linear functions with weights and biases; the likelihood and the upper layer of the prior do not have any learnable parameters.

To generate data, we sample $512$ datapoints from the model where we have set $\vmu_{\vtheta}(\vz_2)=\vz_2$ and $\vsigma_{\vtheta}(\vz_2) = 1$. 

We then train the parameters $\vphi$ and $\vtheta$ in all linear layers using SGD on the IWAE objective til convergence. We then evaluate the gradient variance and gradient signal-to-noise ratio for each estimator. For the proposal parameters $\vphi$ we compare \dregs to the naive score function (labelled as \iwae) and to \stl; for the prior parameters $\vtheta$ we compare \gdregs to \iwae. 

In \cref{fig:toy_2layer} in the main paper we show the average gradient variance and gradient signal-to-noise ratio (SNR). The average is taken over all parameters of either the variational posterior or the prior. The gradient variance and gradient SNR for individual parameters exhibit the same qualitative behaviour.

\subsection{Conditional and unconditional image modelling}
For conditional and unconditional image modelling, we use VAEs with one or multiple stochastic layers, where the generative path is top-down and the inference path is bottom-up, as specified in \cref{eq:vae_models}, which we reproduce here for convenience:
\begin{equation}
    \begin{aligned}
        \qzxc & = q_{\vphi_1}(\vz_1 \given \vx, \vc) \prod_{l=2}^L q_{\vphi_l}(\vz_l \given \vz_{l-1}, \vx, \vc) \\
        \pzc & = p_{\vtheta_L}(\vz_L \given \vc) \prod_{l=1}^{L-1} p_{\vtheta_l}(\vz_l \given \vz_{l+1}, \vc)\\
        \likelzl & = p_{\vlambda}(\vx \given \vz_1, \dots, \vz_L).
    \end{aligned}
     \tag{\ref{eq:vae_models}}
\end{equation}
For conditional image modelling, we predict the bottom half of an image given its top half as in \citet{Tucker2019_dregs}, providing the top half as an additional context input $\vc$ to the prior and variational posterior.
Given the above model structure, $\pa{\valpha}{l} = \vz_{k(l-1)}$ and $\pa{\vbeta}{l} = \vz_{k(l+1)}$.

Each conditional distribution in \cref{eq:vae_models} is given by an MLP of $2$ hidden layers of $300$ \texttt{tanh} units each. If a distribution has multiple inputs, we concatenate them along the feature dimension. The prior and variational posterior are all given by diagonal Gaussian distributions, whereas the likelihood is given by a Bernoulli distribution. The unconditional prior distribution in the uppermost layer $p_{\vtheta_L}(\vz_L)$ is given by a standard Normal distribution, $p_{\vtheta_L}(\vz_L)=\mathcal{N}(0, \mathbb{I})$. All latents $\vz_l$ are $50$ dimensional. 

To avoid overfitting, we use dynamically binarized versions of the datasets. We use the Adam optimizer with default learning rate $3\cdot 10^{-4}$ and default parameters $b_1 = 0.9$, $b_2 = 0.999$, and $\epsilon=10^{-8}$. We use a batch size of $64$ and $K=64$ importance samples for training and evaluation. Note that for testing we report test objective values rather than an estimate of $\log p(\vx)$ by using a large number of importance samples. We do this as we are interested in the relative behaviour of the estimators.

\begin{figure}[htb]
    \includestandalone[mode=buildnew, width=\linewidth]{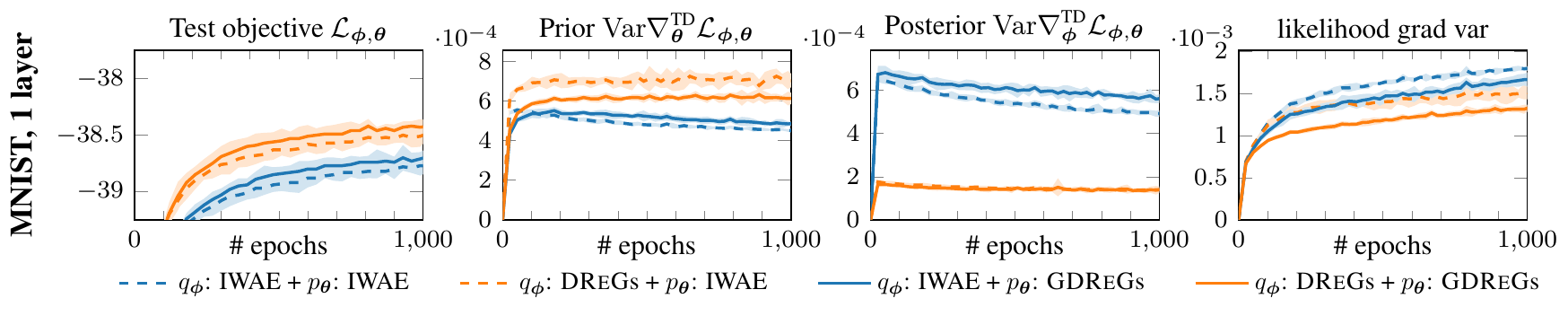}
    \includestandalone[mode=buildnew, width=\linewidth]{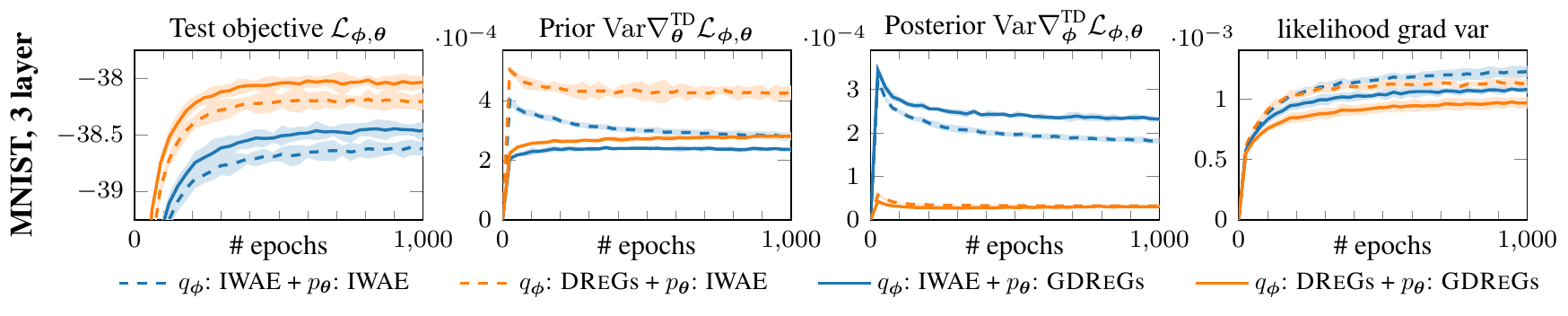}
    \includestandalone[mode=buildnew, width=\linewidth]{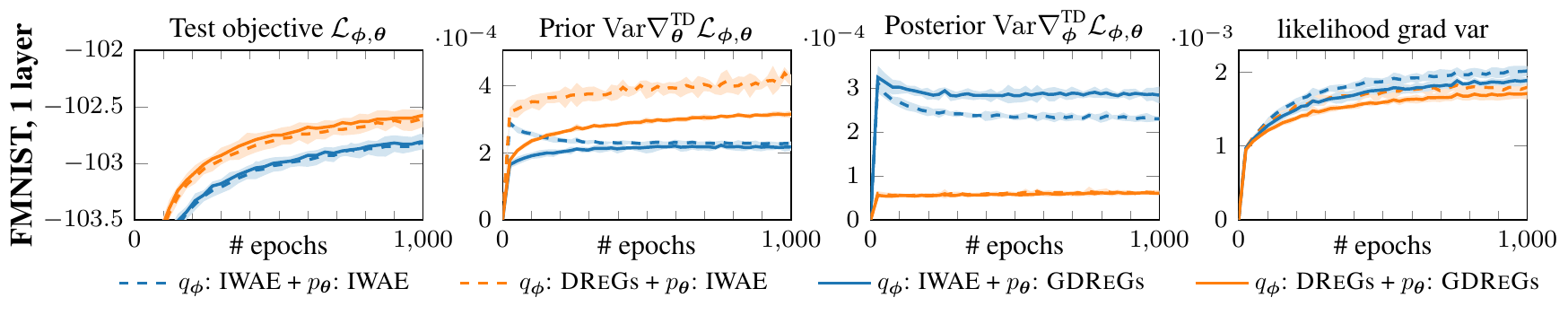}
    \includestandalone[mode=buildnew, width=\linewidth]{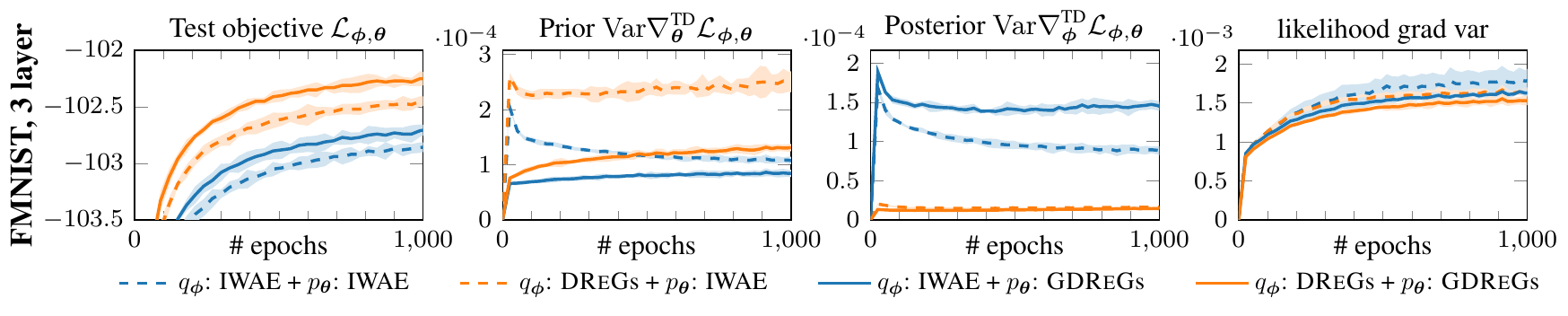}
    \vskip-0.5em
    \caption{Additional results for \emph{conditional} image modelling with VAEs on MNIST and FashionMNIST (FMNIST) with $1$ stochastic layer and $3$ stochastic layers. For each task we show the test objective, the prior gradient variance, the variational posterior gradient variance, and the likelihood gradient variance. Means over $5$ reruns; shaded areas denote $\pm$ $1.96$ standard deviations $\sigma$.}
    \label{app:fig:conditional_image_additional}
\end{figure}

In \cref{app:fig:conditional_image_additional} we provide further plots that show the evolution of the test objective and the gradient variance of the prior, the variational posterior, as well as the likelihood throughout training on \emph{conditional modelling} of MNIST and FashionMNIST (predict bottom half given top half). As in the main paper we find that using \gdregs for the prior instead of the naive estimator (denoted as IWAE) always improves performance on the test objective regardless of the estimator for the variational posterior. We also note that \gdregs always reduces gradient variance for the prior early in training and also typically throughout training, especially for deeper models and when combined with the \dregs estimator for the variational posterior. 

Interestingly, using the \gdregs estimator for the prior leads to an increase in the gradient variance for the variational posterior when we use the naive estimator but similar or even lower posterior gradient variance when combined with the \dregs estimator (third column in \cref{app:fig:conditional_image_additional}). It always leads to a slight improvement in gradient variance for the likelihood parameters (fourth column in \cref{app:fig:conditional_image_additional}). We hypothesize that this is the case because lower gradient variance in for one set of parameters makes it easier to estimate the gradients for another set as a secondary effect.

\begin{figure}[htb]
    \includestandalone[mode=buildnew, width=\linewidth]{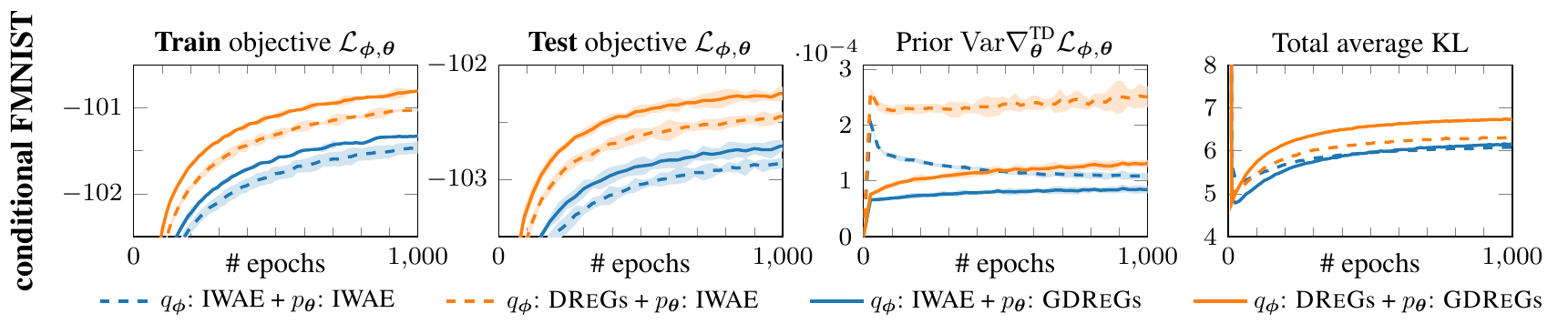}
    \centering\includestandalone[mode=buildnew, width=\linewidth]{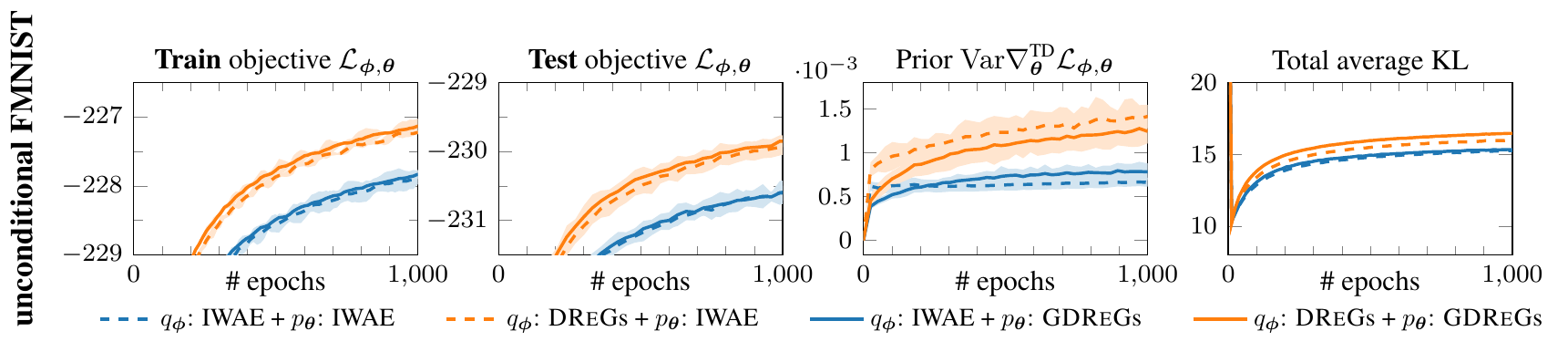}
    \caption{\textbf{Train} objective \textit{(leftmost column)} and total average KL \textit{(rightmost column)} in addition to the test objective and prior gradient variance for conditional and unconditional FashionMNIST on a model with $3$ stochastic layers.}
    \label{app:fig:conditional_vs_unconditional_kl}
\end{figure}

In \cref{app:fig:conditional_vs_unconditional_kl} we show that the training objective behaves qualitatively similar to the test objective in that the applying the \dregs or \gdregs estimator results in improved objective values in the same way, regardless of whether we consider the training or test objective. That is, our hierarchical extension of \dregs results in a better training objective values for both conditional and unconditional tasks. \gdregs is particularly helpful for conditional tasks.

Moreover, in the main text we had hypothesized that \gdregs performs better on conditional image modelling tasks than unconditional tasks because access to the context makes the variational posterior and the prior more similar. In the case of analytically computed cross-entropy in \cref{app:sec:x-entropy}, we derive that the \gdregs estimator outperforms a naive estimator of the score function in terms of gradient variance when the posterior and prior are similar. We hypothesize that this also holds more generally for the IWAE objective, where we cannot compute the gradient variance in closed form but only estimate it empirically.
To investigate this, we computed the total average KL (rightmost column in \cref{app:fig:conditional_vs_unconditional_kl}) over all latent variables and found that it is indeed lower for conditional than unconditional modelling, which indicates that the distributions are closer together in this case.

\subsection{Offline evaluation of the \dregs and \gdregs estimator}
\label{app:sec:offline_eval}

In the image modelling experiments in \cref{sec:exp:generativemodelling} we discussed the gradient variance of the different estimators for the posterior and prior parameters. However, we only analyzed estimators \emph{online} on their respective runs; that is, in \cref{fig:exp:1_2layers_cond_mnist} the prior and posterior gradient variance shown corresponds to the variance of the estimator also used during training. 

Here, we provide an ablation study for conditional image modelling on MNIST with a 2-layer VAE where we evaluate the different estimators offline; that is, for each combination of estimators used for training we also show the gradient variance of the other estimators.

\begin{figure}[htb]
    \centering
    \includestandalone[mode=buildnew, width=\linewidth]{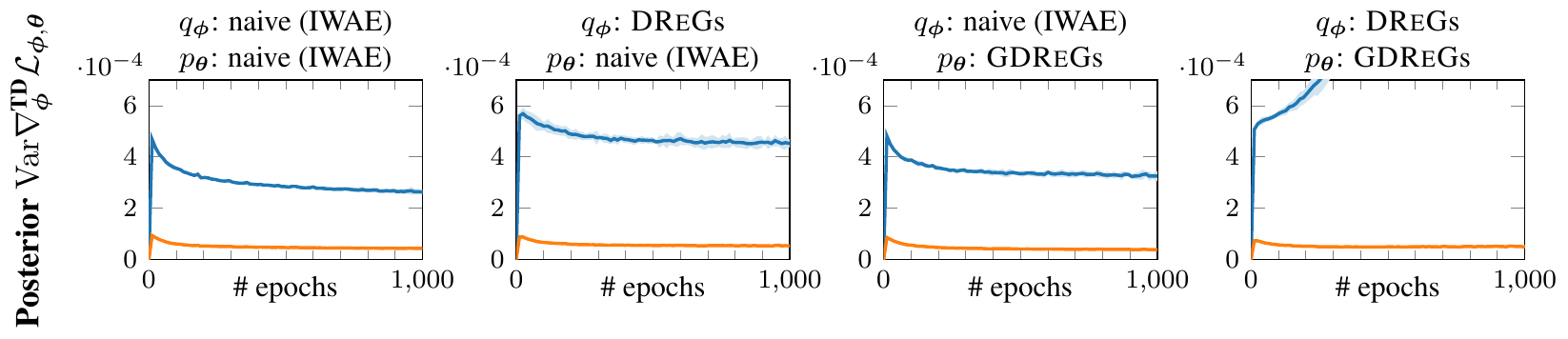}
    \caption[]{Gradient variance of the \textbf{posterior} parameters $\vphi$ as estimated with the naive (IWAE) estimator 
    (\protect\tikz[baseline=-0.5ex,inner sep=0pt]{\protect\draw[very thick, C1] (0,0) -- ++(0.5,0);})
    and the \dregs estimator (\protect\tikz[baseline=-0.5ex,inner sep=0pt]{\protect\draw[very thick, C2] (0,0) -- ++(0.5,0);}) for different combinations of estimators used during training as indicated by the sub-figure title. \\
    For example, in the leftmost plot we compare the posterior gradient variance as estimated by the naive (IWAE) estimator to that of the \dregs estimator on an experiment where we trained both the posterior parameters $\vphi$ as well as the prior parameters $\vtheta$ with the naive (IWAE) estimator.}
    \label{app:fig:offline_posterior}
\end{figure}

In \cref{app:fig:offline_posterior} we consider the gradient variance \wrt the \textbf{variational posterior} parameters $\vphi$ and compare the naive (IWAE) estimator to the \dregs estimator. We find that, regardless which combination of estimators has been used during training, the \dregs estimator \emph{always} results in a better (lower) gradient variance than the naive estimator. 

\begin{figure}[htb]
    \centering
    \includestandalone[mode=buildnew, width=\linewidth]{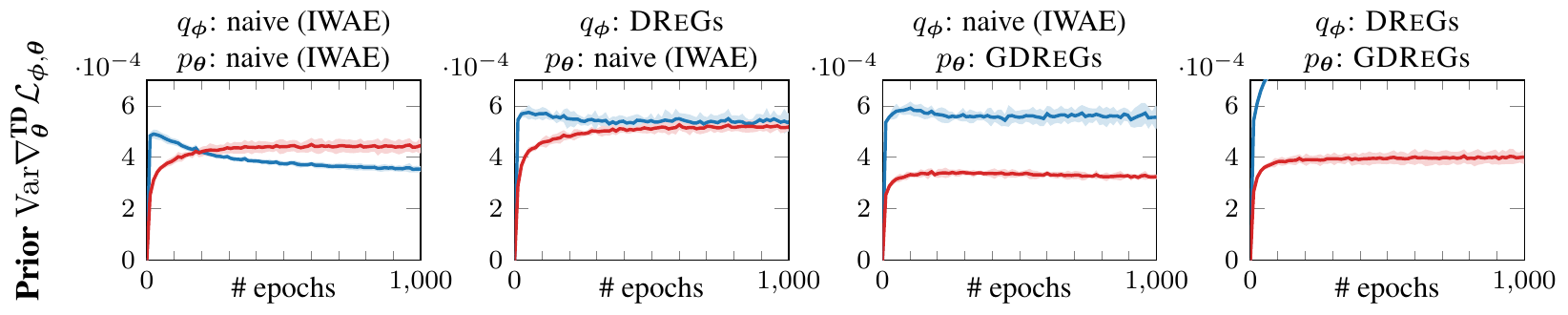}
    \caption[]{Gradient variance of the \textbf{prior} parameters $\vtheta$ as estimated with the naive (IWAE) estimator (\protect\tikz[baseline=-0.5ex,inner sep=0pt]{\protect\draw[thick, C1] (0,0) -- ++(0.5,0);}) and the \gdregs estimator (\protect\tikz[baseline=-0.5ex,inner sep=0pt]{\protect\draw[thick, C4] (0,0) -- ++(0.5,0);}) for different combinations of estimators used during training as indicated by the sub-figure title.\\ 
    For example, in the leftmost plot we compare the prior gradient variance as estimated by the naive (IWAE) estimator to that of the \gdregs estimator on an experiment where we trained both the posterior parameters $\vphi$ as well as the prior parameters $\vtheta$ with the naive (IWAE) estimator.}
    \label{app:fig:offline_prior}
\end{figure}

Similarly, in \cref{app:fig:offline_prior} we consider the gradient variance \wrt the \textbf{prior} parameters $\vtheta$ and compare the naive (IWAE) estimator to the \gdregs estimator. We find that generally the \gdregs gradient estimates have lower variance than the naive (IWAE) estimates. However, when we use the naive estimator for the prior parameters during training, this reduction is smaller and may only be present in the beginning of training. However, consistently, the \gdregs gradient estimates have lower variance when we use the \dregs estimator during training to estimate the variational posterior parameters.

\section{The cross-entropy for Gaussian distributions}
\label{app:sec:x-entropy}

Here, we investigate the properties of the \gdregs estimator compared to the naive estimator in a setting where all quantities of interest can be computed in closed form, the cross-entropy of two Gaussian distributions, $\distqz = \mathcal{N}(\vz; \vmu_q, \vsigma_q)$ and $\distpz = \mathcal{N}(\vz; \vmu_p, \vsigma_p)$.

The negative cross-entropy is given by 
\begin{align}
    \lce = \expect{\vz\sim\distqz}{\log\distpz} & = \tfrac{1}{2}\log(2\pi) + \log\vsigma_p + \frac{\vsigma_q^2 + (\vmu_p-\vmu_q)^2}{2\vsigma_p^2}
\end{align}
Note that in this analytic case, we can compute the gradients without having to sample. However, in the following we want to compare the naive (score function) estimator to the \gdregs estimator.

The naive estimator of this score function is given by: 
\begin{align}
    \naivegrad{\vtheta}{\lce} = \grad{\vtheta}{\log\distpz}; \qquad \vz\sim\distqz
\end{align}
while the \gdregs estimator is given by (see \cref{eq:gdregs:x-entropy}):
\begin{equation}
    \begin{aligned}
    \gdregsgrad{\vtheta}{\lce} = \grad{\vz}{\log \frac{\distqz}{\distpz}}\left.\grad{\vtheta}{\mathcal{T}_p(\vepst; \vtheta)}\right\vert_{\vepst = \mathcal{T}_p^{-1}(\vz, \vtheta)}; \qquad \vz\sim\distqz
\end{aligned}
\tag{\ref{eq:gdregs:x-entropy}}
\end{equation}

\subsection{Gradient variance of the estimators}
In the case under consideration, the parameters $\vtheta$ are given by the mean and variance of the prior, $\vtheta = \{\vmu_p, \vsigma_p\}$, and we can compute both the expectation as well as the variance of these gradient estimators in closed form. 

\paragraph{Naive estimator}
\begin{align}
    \expect{\distqz}{\naivegrad{\vmu_p}{\lce}} & = \frac{\vmu_q-\vmu_p}{\vsigma_p^2} \\
    \variance{\distqz}{\naivegrad{\vmu_p}{\lce}} & = \frac{\vsigma_q^2}{\vsigma_p^4} \label{eq:grad_var_naive_mean}\\
    \expect{\distqz}{\naivegrad{\vsigma_p}{\lce}} & = \frac{\vsigma_q^2-\vsigma_p^2}{\vsigma_p^3} + \frac{\left(\vmu_q - \vmu_p\right)^2}{\vsigma_p^3} \\
    \variance{\distqz}{\naivegrad{\vsigma_p}{\lce}} & = 2\frac{\vsigma_q^4}{\vsigma_p^6} + 4 \vsigma_q^2\frac{\left(\vmu_q-\vmu_p\right)^2 }{\vsigma_p^6}  \label{eq:grad_var_naive_sig}
\end{align}
Note that all operations are element-wise.

\paragraph{The proposed \gdregs estimator}
\begin{align}
    \expect{\distqz}{\gdregsgrad{\vmu_p}{\lce}} & = \frac{\vmu_q-\vmu_p}{\vsigma_p^2} \\
    \variance{\distqz}{\gdregsgrad{\vmu_p}{\lce}} & = \frac{\vsigma_q^2}{\vsigma_p^4} \frac{\left(\vsigma_p^2-\vsigma_q^2\right)^2}{\vsigma_q^4} \label{eq:grad_var_ours_mean}\\
    \expect{\distqz}{\gdregsgrad{\vsigma_p}{\lce}} & = \frac{\vsigma_q^2-\vsigma_p^2}{\vsigma_p^3} + \frac{\left(\vmu_q - \vmu_p\right)^2}{\vsigma_p^3} \\
    \variance{\distqz}{\gdregsgrad{\vsigma_p}{\lce}} & = 2\frac{\left(\vsigma_q^2-\vsigma_p^2\right)^2}{\vsigma_p^6} + \frac{\left(\vsigma_p^2-2\vsigma_q^2\right)^2}{\vsigma_q^2} \frac{\left(\vmu_q-\vmu_p\right)^2 }{\vsigma_p^6} \label{eq:grad_var_ours_sig}
\end{align}
Note that all operations are element-wise.

Both estimators have equal expectation; this is because \gdregs is an unbiased estimator. 

Comparing the variances in \cref{eq:grad_var_naive_mean} and \cref{eq:grad_var_ours_mean} we note that the \gdregs estimator has a lower gradient variance  than the naive estimator for the mean parameters $\vmu_p$ if 
\begin{align}
    \vsigma_p^2 \leq 2 \vsigma_q^2. 
\end{align}
Similarly, comparing \cref{eq:grad_var_naive_sig} and \cref{eq:grad_var_ours_sig}, we find that the \gdregs estimator has lower variance than the naive estimator for the variance parameters $\vsigma_p$ if
\begin{align}
    \vsigma_p^2 \leq 4\vsigma_q^2\left(1-\frac{\vsigma_q^2}{(\vmu_p-\vmu_q)^2+2\vsigma_q^2}\right)
\end{align}
In the case of $\vmu_p = \vmu_q$, this also reduces to $\vsigma_p^2 \leq 2 \vsigma_q^2$.

Thus, we expect the \gdregs estimator to perform better than the naive estimator when \distpz and \distqz are close together.

\subsection{Constructing the optimal control variate}

Because the GDReGs estimators and the naive estimators have the same expectation, we can build a control variate out of their difference:
\begin{align}
    \left[\hat{\nabla}_{\vtheta}^\text{naive} + \alpha( \hat{\nabla}_{\vtheta}^\text{GDReG} - \hat{\nabla}_{\vtheta}^\text{naive})\right] \lce.
\end{align}
We can then compute its optimal strength $\alpha^\ast$, by minimizing its variance,
\begin{align}
    \alpha^\ast = \arg\min_{\alpha} \variance{\distqz}{\left[\hat{\nabla}_{\vtheta}^\text{naive} + \alpha( \hat{\nabla}_{\vtheta}^\text{GDReG} - \hat{\nabla}_{\vtheta}^\text{naive})\right]\lce}.
\end{align}

We find that:
\begin{align}
    \alpha_{\vmu}^\ast & = \frac{\vsigma_q^2}{\vsigma_p^2} \\
    \alpha_{\vsigma}^\ast & = \frac{2\vsigma_q^2}{\vsigma_p^2} \frac{(\vmu_p-\vmu_q)^2 + \vsigma_q^2}{(\vmu_p-\vmu_q)^2 + 2\vsigma_q^2} \\
    \variance{\distqz}{\left[\hat{\nabla}_{\vmu_p}^\text{naive} + \alpha^\ast_{\vmu}( \hat{\nabla}_{\vmu_p}^\text{GDReG} - \hat{\nabla}_{\vmu_p}^\text{naive})\right]\lce} & = 0\\
    \variance{\distqz}{\left[\hat{\nabla}_{\vsigma_p}^\text{naive} + \alpha^\ast_{\vsigma}( \hat{\nabla}_{\vsigma_p}^\text{GDReG} - \hat{\nabla}_{\vsigma_p}^\text{naive})\right]\lce} &= \frac{2\vsigma_q^4}{\vsigma_p^6}\frac{(\vmu_q - \vmu_p)^2}{(\vmu_q - \vmu_p)^2 + 2\vsigma_q^2}
\end{align}

Note that the expression for the optimal strength has different form for the mean $\vmu_p$ and variance $\vsigma_p$ parameters. Moreover, note that the analytic estimator has zero gradient variance whereas our estimator with control variate still has non-zero gradient variance for the variance parameters $\vsigma_p$.

This optimal estimator holds for a single layer VAE, where both the variational posterior as well as the prior are Gaussian. In a hierarchical model where both the prior and the posterior are factorized top-down, the same derivation holds for the lowest stochastic layer (similar to how semi-analytic approximations for the conditional KL can be derived in this case). Unfortunately, in this case the expectations cannot be computed in closed form anymore.

\end{document}